\pdfoutput=1

\documentclass[11pt]{article}

\usepackage[final]{acl}

\usepackage{times}
 \usepackage{multirow}
 \usepackage{booktabs}
 \usepackage{amssymb}
\usepackage{makecell}

\usepackage{amsmath,comment}
\usepackage{subcaption}
\usepackage[T1]{fontenc}

\usepackage[utf8]{inputenc}

\usepackage{microtype}

\usepackage{inconsolata}

\usepackage{graphicx}

\title{How Good are LLM-based Rerankers?
An Empirical Analysis of State-of-the-Art Reranking Models }

\author{
    \textbf{Abdelrahman Abdallah, Bhawna Piryani, Jamshid Mozafari, Mohammed Ali, Adam Jatowt} \\
    University of Innsbruck \\
    \texttt{\{abdelrahman.abdallah, jamshid.mozafari, bhawna.piryani,} \\
    \texttt{ 
mohammed.ali, adam.jatowt\}@uibk.ac.at}
}

\begin{document}
\maketitle
\begin{abstract}
In this work, we present a systematic and comprehensive empirical evaluation of state-of-the-art reranking methods, encompassing large language model (LLM)-based, lightweight contextual, and zero-shot approaches, with respect to their performance in information retrieval tasks.
We evaluate in total 22 methods, including 40 variants (depending on used LLM) across several established benchmarks, including TREC DL19, DL20, and BEIR, as well as a novel dataset designed to test queries unseen by pretrained models.
Our primary goal is to determine, through controlled and fair comparisons, whether a performance disparity exists between LLM-based rerankers and their lightweight counterparts, particularly on novel queries, and to elucidate the underlying causes of any observed differences.
To disentangle confounding factors, we analyze the effects of training data overlap, model architecture, and computational efficiency on reranking performance.
Our findings indicate that while LLM-based rerankers demonstrate superior performance on familiar queries, their generalization ability to novel queries varies, with lightweight models offering comparable efficiency.
We further identify that the novelty of queries significantly impacts reranking effectiveness, highlighting limitations in existing approaches \footnote{\url{https://github.com/DataScienceUIBK/llm-reranking-generalization-study}
}.

\end{abstract}

\section{Introduction}

\begin{figure}[htbp]
    \centering
    \includegraphics[width=0.5\textwidth]{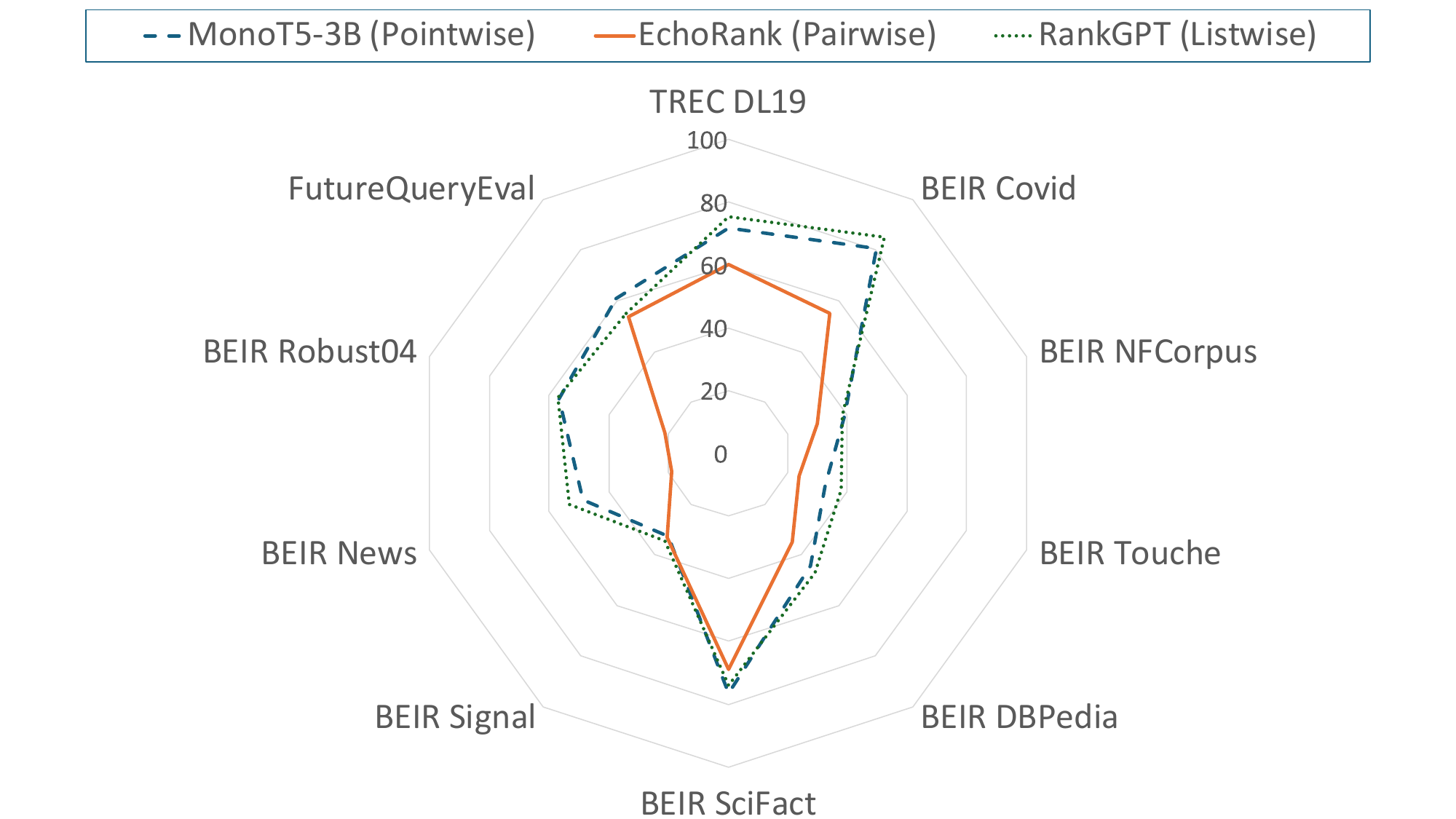}
\caption{Radar chart comparing nDCG@10 performance of top pointwise (MonoT5-3B), pairwise (PRP-FLAN-UL2), and listwise (RankGPT-gpt-4) reranking methods across TREC DL19, all BEIR datasets, and FutureQueryEval. DL20 is excluded to maintain chart readability given the large number of datasets displayed.}
\label{fig:radar_chart_all_beir}
\end{figure}
Text reranking, the task of refining retrieved documents to optimize relevance to a user query, is crucial for information retrieval (IR) systems, including web search \cite{yasser2018re}, open-domain question answering \cite{chen2017reading,gruber2024complextempqa}, and retrieval-augmented generation (RAG) \cite{lewis2020retrieval,abdallah2025tempretriever}. Transformer-based models and large language models (LLMs), such as BERT \cite{devlin2019bert} and GPT-4 \cite{achiam2023gpt}, have advanced reranking with strong contextual understanding and zero-shot capabilities \cite{kojima2022large}. 

However, the reliance on large training corpora raises concerns about generalization ability with respect to novel queries
unseen during pretraining \cite{sun2023chatgpt}. Despite the emergence of LLM-based \cite{mao2024rafe} and lightweight rerankers like ColBERT \cite{khattab2020colbert}, claims of superior performance often lack rigorous evidence due to data contamination in standard datasets. As noted by \citet{yu2022generate,abdallah2023generator,mozafari2024exploring}, existing benchmark questions are typically gathered years ago, which raises the issue that existing LLMs already possess knowledge of these questions. This contamination risk is acknowledged even by model developers, with OpenAI~\cite{achiam2023gpt} noting the potential risk of contamination of the existing benchmark test set.  Recent advances have introduced additional reranking paradigms beyond the traditional pointwise, pairwise, and listwise approaches. Setwise reranking \cite{zhuang2024setwise} processes documents in sets rather than individually or in pairs, offering a middle ground between pairwise and listwise complexity. TourRank \cite{chen2025tourrank} introduces a tournament-style ranking approach that recursively compares document subsets. Additionally, specialized LLM-based rerankers like DynRank~\cite{abdallah-etal-2025-dynrank}, ASRank~\cite{abdallah2025asrank}, and RankLLaMA~\cite{ma2024fine}.

\begin{figure*}[htbp]
    \centering

    \begin{subfigure}[b]{0.30\textwidth}
        \includegraphics[width=\linewidth]{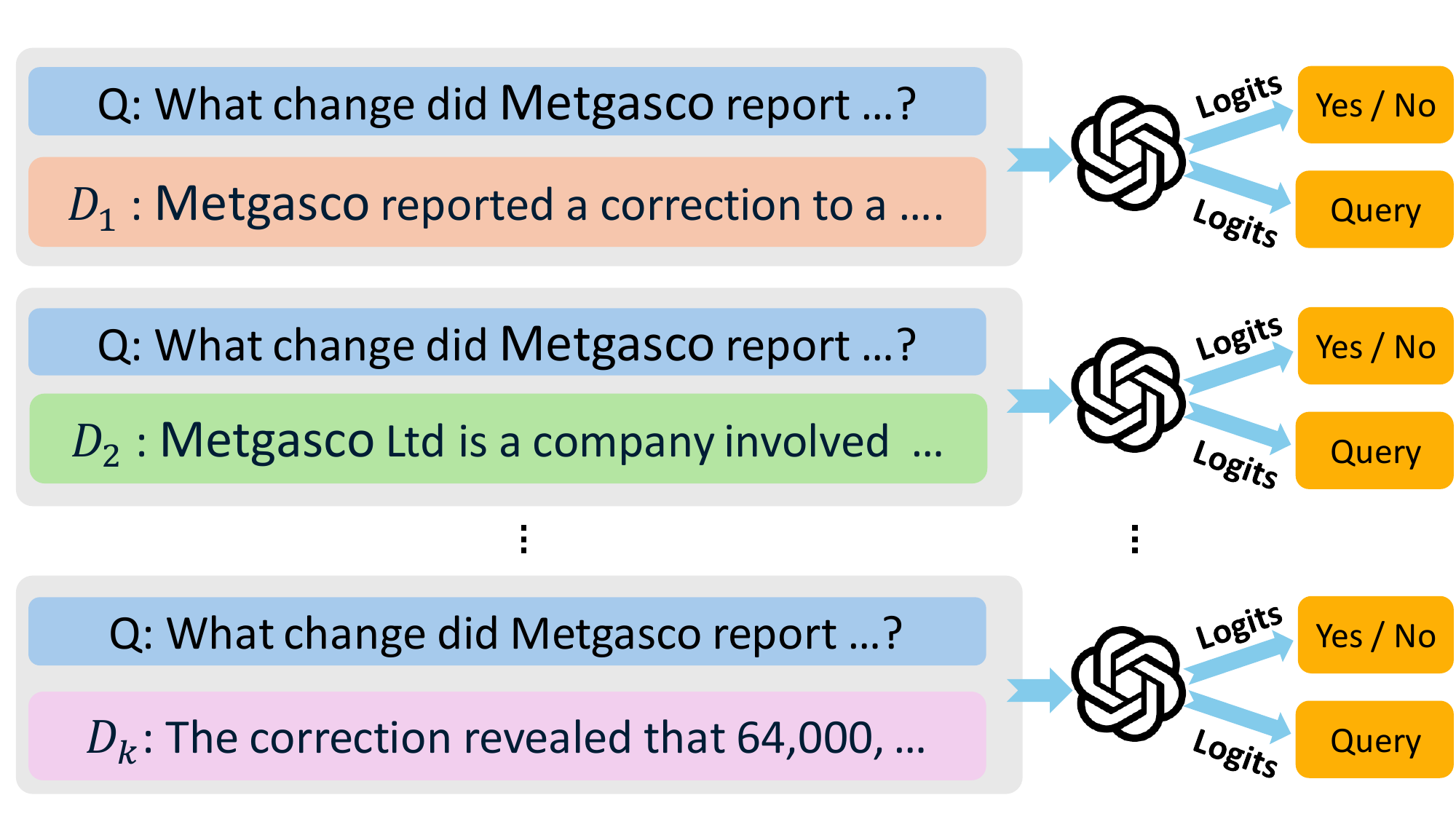}
        \caption{Pointwise approach}
        \label{fig:Pointwiseapproach}
    \end{subfigure}
    \hfill
    \begin{subfigure}[b]{0.3\textwidth}
        \includegraphics[width=\linewidth]{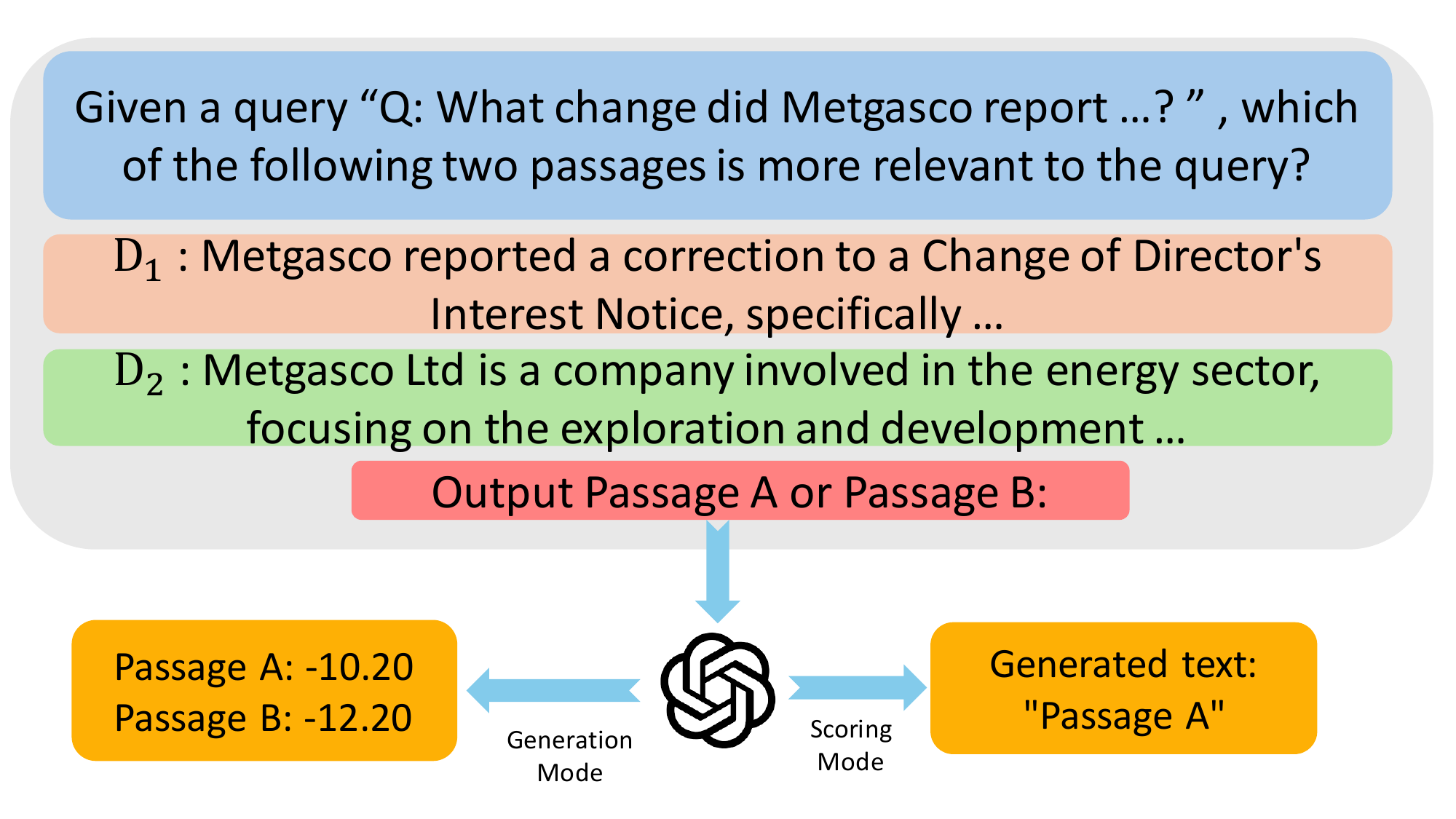}
        \caption{Pairwise approach}
        \label{fig:Pairwiseapproach}
    \end{subfigure}
    \hfill
    \begin{subfigure}[b]{0.3\textwidth}
        \includegraphics[width=\linewidth]{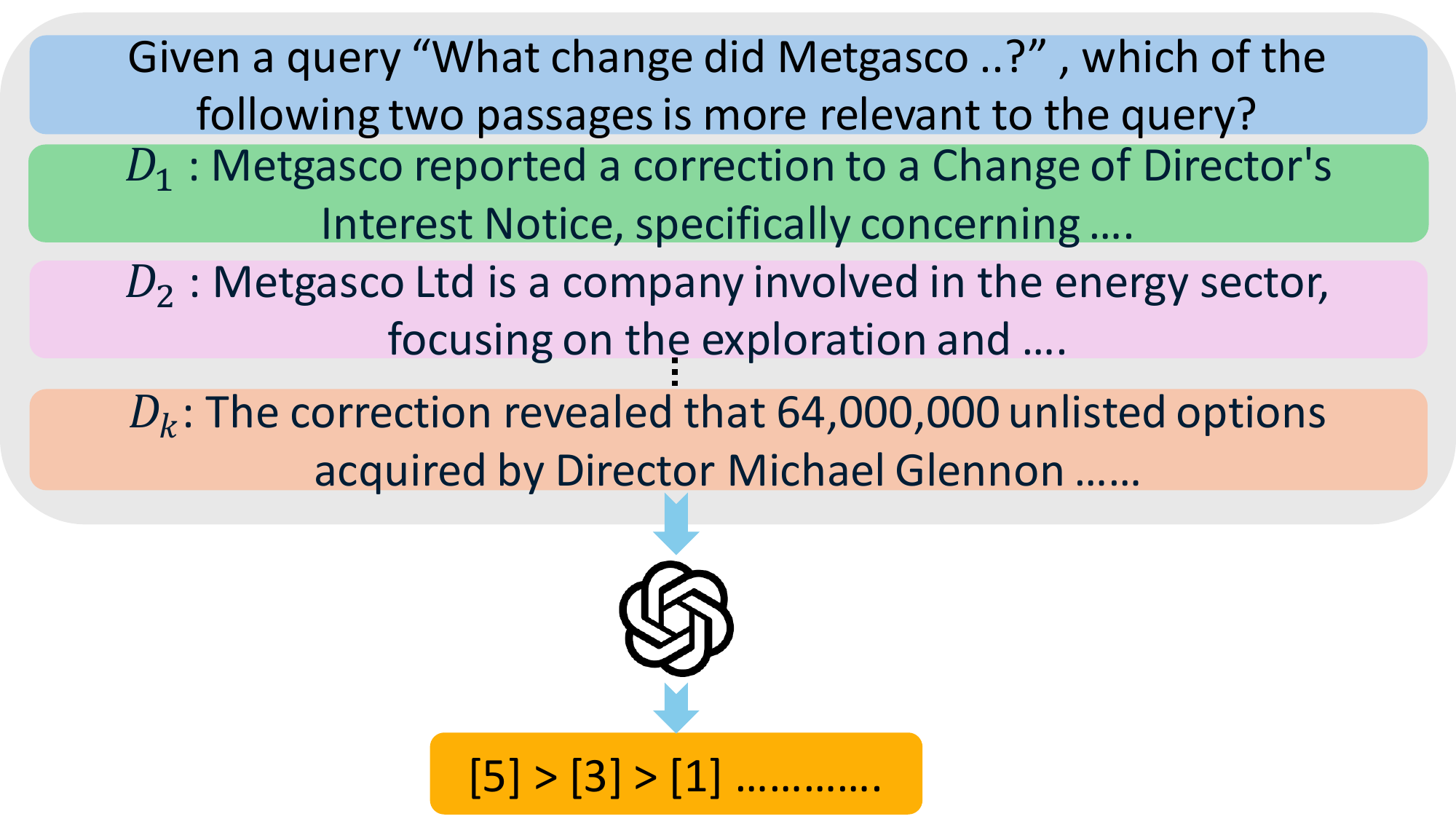}
        \caption{Listwise approach}
        \label{fig:Listwiseapproach}
    \end{subfigure}

    \caption{Illustration of reranking approaches: (a) Pointwise approach, scoring each query-document pair independently; (b) Pairwise approach, comparing pairs of documents to determine relative relevance; (c) Listwise approach, processing multiple documents simultaneously to generate a reordered list.}
    \label{fig:three_subfigs}
\end{figure*}
Current reranking models are typically benchmarked on standard datasets like TREC DL19, DL20 \cite{craswell2020overview}, and BEIR \cite{thakur2021beir} containing well-studied queries \cite{khattab2020colbert, zhuang2023beyond}. We hypothesize that reranker performance varies with novel queries, affecting efficiency and robustness. We introduce FutureQueryEval, a dataset with queries absent from LLM training until May 2025, for fair evaluation. Figure~\ref{fig:radar_chart_all_beir} compares pointwise, pairwise, and listwise methods across TREC DL19, BEIR, and FutureQueryEval, showing generalisation challenges. Our analysis investigates these factors by comparing state-of-the-art rerankers, including LLM-based, lightweight, and zero-shot models, on both standard benchmarks and our custom dataset. We also explore the interplay of model architecture, training data overlap, and computational efficiency, shedding light on the trade-offs that influence reranking performance.

\vspace{1.6mm} \noindent \textbf{Contributions}. \textbf{1)} We introduce a novel dataset with queries absent from LLM training data until May 2025, enabling unbiased evaluation of reranking methods. \textbf{2)} We systematically compare LLM-based, lightweight, and zero-shot reranking approaches on TREC DL19, DL20, BEIR, and our custom dataset. \textbf{3) }We analyze key factors affecting reranking performance, including generalization to novel queries, computational efficiency, and model architecture. \textbf{4)} Our findings provide actionable insights into the robustness and scalability of reranking methods, guiding the development of future IR systems.

\section{Related Work}

Large language models (LLMs) have transformed information retrieval (IR) by enabling semantic understanding and zero-shot ranking capabilities. Retrieval-Augmented Generation (RAG) \cite{jiang2024longrag} integrates retrieval with LLM generation to enhance response quality. Reranking \cite{abdallah2025asrank} refines retrieved documents, prioritizing relevance to improve RAG outcomes and reduce LLM hallucinations. \citet{gao2023retrieval} highlight reranking’s role in evolving RAG frameworks, boosting accuracy in tasks like question answering. \citet{zhao2024retrieval} note that reranking supports multimodal RAG, mitigating data leakage by refining diverse data types. \citet{Yu_2025} propose metrics like relevance to evaluate reranking’s impact, emphasizing robust strategies for reliable RAG performance. Fairness in LLM-based ranking is critical for equitable applications. \citet{wang2024large} find that LLMs like GPT and Llama2 under-represent groups on the TREC Fair Ranking dataset, with exposure disparities up to 15\%. Traditional methods like FA*IR \cite{zehlike2017fa} and exposure metrics \cite{singh2018fairness} struggle with LLMs’ opaque decisions, underscoring the need for fairness-aware ranking approaches.

\vspace{-2mm}
\section{Reranking Approaches}
Reranking in information retrieval (IR) refines an initial set of retrieved documents to optimize their relevance to a query, a critical step in applications like web search and question answering. With the rise of pretrained language models (PLMs) and large language models (LLMs), reranking methods have evolved into three primary categories: pointwise, pairwise, and listwise. These approaches differ in how they score and order documents, balancing effectiveness, efficiency, and generalization. This section presents an overview of key types of reranking methods, detailing their methodologies, and key implementations.

\subsection{Pointwise Reranking}

Pointwise reranking assigns independent relevance scores to query-document pairs, computed by classification or regression, and sorts documents by scores. With $O(n)$ complexity, this approach is efficient for large-scale use yet it suffers from the lack of explicit consideration of inter-document dependencies, thus preventing relative relevance modeling.

Transformer-based models have advanced pointwise reranking. ~\citet{nogueira2019passage} proposed \textbf{monoBERT}, using BERT for binary classification, concatenating query and document to output a relevance score via the [CLS] token. This method performs strongly on MS MARCO and TREC in multi-stage pipelines. ~\citet{nogueira2020document} introduced \textbf{MonoT5}, adapting T5 with a prefix (e.g., ``Query: q Document: d Relevant:'') to predict ``true'' or ``false'' for relevance, using the ``true'' token’s probability as the score. MonoT5 surpasses monoBERT on MS MARCO and excels in zero-shot settings on TREC 2004 Robust Track. ~\citet{zhuang2023rankt5} developed \textbf{RankT5}, directly outputting numerical scores with encoder-decoder or encoder-only architectures, fine-tuned with ranking losses for efficiency. ~\citet{laitz2024inranker} presented \textbf{InRanker}, distilling MonoT5-3B into smaller models (60M, 220M parameters) for zero-shot reranking, trained on MS MARCO and synthetic BEIR labels via InPars, achieving 50$\times$ size reduction ~\cite{bonifacio2022inpars}. Finally, UPR~\cite{sachan2022improving}, ASRANK~\cite{abdallah2025asrank} and DynRank ~\cite{abdallah-etal-2025-dynrank} explored unsupervised methods scoring relevance as the query’s likelihood given a passage using a pretrained model with a prompt. This approach generalizes to new domains without the need for task-specific training.

\subsection{Pairwise Reranking}

Pairwise reranking compares document pairs to determine relative relevance, aggregating results to form a ranking. With $O(n^2)$ complexity for all-pair comparisons, optimized variants achieve $O(n \log n)$ or $O(n)$. It excels in precise differentiation but faces scalability and transitivity challenges.

LLM-based pairwise methods leverage large language models for effective reranking. ~\citet{qin2023large} introduced Pairwise Ranking Prompting (PRP), prompting an LLM (e.g., FlanUL2) to select the more relevant document from a query-document pair, using scoring APIs for reliability ~\cite{qin2023large}. Variants include PRP-Allpair (win-ratio aggregation), PRP-Sorting (Heapsort, $O(n \log n)$), and PRP-Sliding-K (sliding window, $O(n)$), with FlanUL2 outperforming InstructGPT by 10\% in NDCG@10 on TREC-DL2019/2020. Similarly, ~\citet{jiang2023llm} proposed PAIRRANKER within the LLM-BLENDER framework, encoding query and two LLM outputs with a cross-attention Transformer (e.g., RoBERTa) to compute confidence scores, aggregated via MaxLogits or bubble sort ($O(n)$) ~\cite{jiang2023llm}. Evaluated on MixInstruct, PAIRRANKER achieves a 68.59\% top-3 ranking rate, surpassing pointwise baselines. ~\citet{rashid2024ecorank} developed \textbf{EcoRank}, a budget-conscious two-stage pipeline ~\cite{rashid2024ecorank}. A costly LLM (e.g., FlanT5-XL) filters passages via pointwise classification, followed by pairwise comparisons using a cheaper LLM (e.g., FlanT5-L) with a sliding window, balancing cost and quality.

\begin{table*}[ht]
\scriptsize
\centering
\caption{Comparison of pointwise, pairwise, and listwise reranking challenges. $n$ is the number of documents per query. $O(n)$ for listwise assumes sliding window or tournament sort, as full permutation is impractical. }
\begin{tabular}{c|c|c|c|c|c}
\toprule
Method & \# of LLM API Calls & Generation API & Scoring API & Require Calibration & Sensitivity to Input Order \\
\midrule
Pointwise & $O(n)$ & No & Yes & Yes & Low \\
Pairwise & $O(n^2), O(n\log n), O(n)$ & Yes & Yes & No & Medium \\
Listwise & $O(n)$ & Yes & No & No & High \\
\bottomrule
\end{tabular}
\label{tbl:comparison}
\end{table*}
\subsection{Listwise Reranking}

Listwise reranking processes a query and multiple documents simultaneously, outputting a reordered list by capturing inter-document relationships. With $O(n)$ complexity, it offers superior accuracy over pointwise and pairwise methods but faces challenges with long input contexts and positional biases when using large language models (LLMs).

LLM-based listwise methods use prompting for zero-shot reranking. ~\citet{sun2023chatgpt} introduced \textbf{RankGPT}, using ChatGPT or GPT-4 to generate passage identifier permutations (e.g., ``[2] > [3]'') with a sliding window to handle token limits. ~\citet{ma2023zero} proposed \textbf{LRL}, employing GPT-3 to reorder passages via a simple prompt and sliding window strategy. ~\citet{pradeep2023rankvicuna} developed \textbf{RankVicuna}, a 7B-parameter LLM distilled from RankGPT3.5, using shuffled inputs for robustness. ~\citet{pradeep2023rankzephyr} presented \textbf{RankZephyr}, a 7B LLM fine-tuned on RankGPT4 data with multiple reranking passes for enhanced performance.

Efficiency-focused listwise methods optimize latency and context handling. ~\citet{yoon2024listt5} introduced \textbf{ListT5}, using T5’s Fusion-in-Decoder to encode query-passage pairs and decode sorted identifiers with $O(n + k \log n)$ complexity. 

~\citet{reddy2024first} proposed \textbf{FIRST}, generating rankings from first-token logits, decreasing latency by 50\% using joint loss training ~\cite{reddy2024first}. ~\citet{liu2025leveraging} presented \textbf{PE-Rank}, using passage embeddings and dynamic decoding to reduce latency by 4.5$\times$ ~\cite{liu2025leveraging}. ~\citet{chen2024attention} developed \textbf{ICR}, leveraging LLM attention weights for $O(1)$ reranking, outperforming RankGPT on TREC and BEIR ~\cite{chen2024attention}.

\section{Challenges in Reranking with LLM}
\label{sec:challenges}

Reranking refines retrieved documents to match queries in information retrieval (IR). LLMs enable zero-shot reranking but face challenges due to their general-purpose design, hindering performance compared to fine-tuned rankers. Issues include computational complexity, API reliance, and prediction inconsistencies across pointwise, pairwise, and listwise methods.

\paragraph{Pointwise Reranking Challenges}
\label{sec:pointwise}

Pointwise methods score query-document pairs independently as classification or regression, sorting by scores with $O(n)$ complexity. A prompt (Figure~\ref{fig:Pointwiseapproach}) yields a relevance judgment (e.g., ``Yes'' or ``No''), with scores defined as:
\begin{equation}
s_i =
\begin{cases}
1 + p(\text{Yes}), & \text{if output is Yes} \\
1 - p(\text{No}), & \text{if output is No}
\end{cases}
\label{eq:pointwise_score}
\end{equation}
where $p(\text{Yes})$ and $p(\text{No})$ are scoring probabilities. Challenges include inconsistent score calibration across prompts, unnecessary for ranking~\cite{desai2020calibration}, and reliance on scoring APIs, limiting compatibility with generation-only LLMs like GPT-4 ~\cite{laitz2024inranker,sachan2022improving}.

\paragraph{Pairwise Reranking Challenges}
\label{sec:pairwise}

Pairwise methods compare document pairs for relative relevance, aggregating results with complexity from $O(n^2)$ to $O(n)$. A prompt (e.g., PRP) selects the more relevant document, scoring $d_i$ as:
\begin{equation}
s_i = \sum_{j \neq i} \left( \mathbb{I}_{d_i > d_j} + 0.5 \cdot \mathbb{I}_{d_i = d_j} \right),
\label{eq:pairwise_score}
\end{equation}
where $\mathbb{I}_{d_i > d_j}$ indicates preference ~\cite{qin2023large}. High complexity for large $n$ and inconsistent judgments for subtle differences ~\cite{jiang2023llm} pose challenges, amplified by sensitivity to initial retrieval quality.

\begin{table*}[!t]
\centering
\small
\setlength\tabcolsep{2pt}
\resizebox{0.8\linewidth}{!}{%

\begin{tabular}{l l cc|cccccccc}
\toprule
\textbf{Method} & \textbf{Model} & \textbf{DL19} & \textbf{DL20} & \textbf{Covid} & \textbf{NFCorpus} & \textbf{Touche} & \textbf{DBPedia} & \textbf{SciFact} & \textbf{Signal} & \textbf{News} & \textbf{Robust04} \\
\midrule
\multirow{7}{*}{UPR} & T5-Small & 49.94 & 47.07 & 62.81 & 30.46 & 21.54 & 31.98 & 61.68 & 29.71 & 27.41 & 33.14 \\
 & T5-Base & 55.12 & 53.94 & 64.12 & 31.12 & 19.42 & 34.28 & 65.56 & 30.50 & 27.32 & 33.40 \\
 & T5-Large & 58.33 & 56.05 & 68.69 & 31.71 & 18.94 & 35.35 & 65.44 & 32.80 & 25.25 & 34.48 \\
 
 & T0-3B & 60.18 & 59.55 & 68.83 & 33.48 & 23.97 & 34.41 & 71.21 & 33.02 & 39.10 & 41.74 \\
 & FLAN-T5-XL & 53.85 &  56.02 &  68.11 &  35.04  & 19.69  & 30.91  & 72.69  & 31.91 &  43.11 &  42.43  \\
 & GPT2 & 50.71 & 44.98 & 62.31 & 31.87 & 18.24 & 29.11 & 64.95 & 32.31 & 31.31 & 34.27 \\
 & GPT2-medium & 51.70 & 49.62 & 63.72 & 31.93 & 16.59 & 30.11 & 65.11 & 32.02 & 32.08 & 35.31 \\
 & GPT2-large & 52.48 & 0.467 & 63.23 & 33.62 & 16.72 & 30.76 & 32.49 & 66.59 & 30.85&  34.69\\
 \midrule

\multirow{5}{*}{FlashRank} & TinyBERT & 67.68 & 60.65 & 61.48 & 32.85 & 32.72 & 36.04 & 64.08 & 31.85 & 37.53 & 41.37 \\
 & MiniLM\footnote{\url{https://huggingface.co/cross-encoder/ms-marco-MiniLM-L12-v2}} & 70.80 & 66.27 & 69.06 & 33.02 & 34.77 & 42.77 & 66.28 & 33.62 & 44.54 & 47.18 \\
 & MultiBERT & 31.29 & 28.47 & 39.62 & 26.84 & 25.17 & 17.56 & 29.14 & 17.39 & 23.13 & 23.09 \\
 & T5-Flan & 21.79 & 17.02 & 38.61 & 18.11 & 8.23 & 7.77 & 8.29 & 6.57 & 12.06 & 15.40 \\
 & MiniLM\footnote{Fine-tuned on Amazon ESCI dataset} & 70.40 & 65.60 & 69.66 & 32.80 & 34.61 & 39.75 & 59.14 & 28.10 & 41.44 & 46.09 \\
 \midrule
\multirow{6}{*}{MonoT5} & Base & 70.81 & 67.21 & 72.24 & 34.81 & 38.24 & 42.01 & 73.14 & 30.47 & 45.12 & 51.24 \\
 & Base-10k & 71.38 & 66.31 & 74.61 & 35.69 & 37.86 & 42.09 & 73.39 & 32.14 & 46.09 & 51.69 \\
 & Large & 72.12 & 67.11 & 77.38 & 36.91 & 38.31 & 41.55 & 73.67 & 33.17 & 47.54 & 56.12 \\
 & Large-10k & 72.12 & 67.11 & 77.38 & 36.91 & 38.31 & 41.55 & 73.67 & 33.17 & 47.54 & 56.12 \\

 & mT5-Base & 70.81 & 64.77 & 73.77 & 34.36 & 35.62 & 40.11 & 71.17 & 29.79 & 45.34 & 48.99 \\

& 3B & 71.83 & 68.89 & 80.71 &  38.97 & 32.41 &44.45 &  76.57 & 32.55& 48.49 & 56.71\\
 \midrule
\multirow{3}{*}{RankT5} & T5-base & 72.13 & 67.91 & 75.63 & 34.99 & 41.24 & 42.39 & 73.37 & 30.86 & 44.07 & 52.19 \\
 & T5-large & 72.82 & 67.37 & 75.45 & 36.27 & 39.34 & 42.90 & 74.84 & 32.53 & 46.81 & 54.48 \\
 & T5-3b & 71.09 & 68.67 & 80.43 & 37.43 & 40.41 & 42.69 & 76.58 & 31.77 & 48.05 & 55.91 \\
  \midrule

\multirow{3}{*}{Inranker} & Inranker-small & 69.81 &61. 68 & 77.75 & 35.47 &  28.83 &  44.51  &74.90 & 29.37 & 46.29 & 50.91\\
 & Inranker-base  & 71.84 & 66.30& 79.84  &  36.58 & 28.97  & 46.50& 76.18 & 30.46 & 47.88 &  54.27\\
 & Inranker-3b  & 72.71 & 67.09 & 81.75  & 38.25  &29.24   &47.62& 78.31 &  32.20 & 49.63 &  62.47 \\
  \midrule
\multirow{10}{*}{Transformer Ranker} & mxbai-rerank-xsmall & 68.95 & 63.11 & 80.80 & 34.44 & 39.44 & 42.5 & 68.73 & 29.40 & 53.00 & 53.87 \\

 & mxbai-rerank-base & 72.49 & 67.15 & 84.00 & 35.64 & 34.32 & 42.50 & 72.33 & 30.20 & 51.92 & 55.59 \\
 
 & mxbai-rerank-large & 71.53 & 69.45 & 85.33 & 37.08 & 36.90 & 44.51 & 75.10 & 31.90 & 51.90 & 58.67 \\
 & bge-reranker-base & 71.17 & 66.54 & 67.50 & 31.10 & 34.30 & 41.50 & 70.60 & 28.40 & 39.50 & 42.90 \\
 & bge-reranker-large & 72.16 & 66.16 & 74.30 & 34.80 & 35.60 & 43.70 & 74.10 & 30.50 & 43.40 & 49.90 \\
 & bge-reranker-v2-m3 & 72.19 & 66.98 & 74.79 & 33.84 & 39.85 & 41.93 & 73.48 & 31.36 & 45.84 & 48.44 \\

 & bce-reranker-base & 70.45 & 64.13 & 67.59 & 33.90 & 27.50 & 38.14 & 70.15 & 27.31 & 40.48 & 48.13 \\
   
 & jina-reranker-tiny & 70.43 & 65.31 &  77.15 & 37.24 & 31.04 & 42.14 & 73.42 & 32.25 & 42.27 & 47.41\\

 & jina-reranker-turbo & 70.35 & 63.62 & 77.97 & 37.29 & 30.80 & 41.75 & 74.53 & 28.46 & 42.79 & 44.19 \\

  \midrule

\multirow{1}{*}{Splade Reranker} & Splade Cocondenser & 71.47 & 66.18 & 68.87 & 34.95 & 37.96 & 41.25 & 68.72 & 32.27 & 43.28 & 47.51 \\
 \midrule
\multirow{12}{*}{Sentence Transformer Reranker} & all-MiniLM & 63.84 & 60.40 & 70.83 & 33.10 & 29.23 & 34.87 & 65.63 & 28.50 & 45.42 & 46.03 \\
 & GTR-T5-base & 68.09 & 62.40 & 70.10 & 32.02 & 32.70 & 36.20 & 60.23 & 30.79 & 43.24 & 45.38 \\
 & GTR-T5-large & 67.23 & 63.33 & 69.50 & 33.03 & 32.84 & 38.20 & 62.41 & 31.19 & 44.32 & 46.98 \\
 & GTR-T5-xl & 67.55 & 64.51 & 69.63 & 33.39 & 34.28 & 38.76 & 63.65 & 31.10 & 45.73 & 47.95 \\
 & GTR-T5-xxl & 68.53 & 64.07 & 72.70 & 34.02 & 36.77 & 39.90 & 65.62 & 31.37 & 47.01 & 49.67 \\
 & sentence-T5-base & 51.15 & 49.37 & 66.02 & 30.17 & 24.63 & 33.67 & 47.29 & 29.78 & 41.71 & 48.24 \\
 & sentence-T5-xl & 54.95 & 53.22 & 67.01 & 31.72 & 29.78 & 36.38 & 50.73 & 31.22 & 43.58 & 48.33 \\
 & sentence-T5-xxl & 60.61 & 58.37 & 72.55 & 34.76 & 30.88 & 40.52 & 60.23 & 31.05 & 49.51 & 52.45 \\
 & sentence-T5-large & 55.36 & 54.20 & 63.57 & 30.23 & 28.08 & 31.89 & 47.40 & 30.56 & 42.94 & 47.06 \\
 & msmarco-bert-co-condensor & 56.34 & 53.50 & 62.20 & 28.38 & 20.12 & 31.93 & 53.04 & 31.16 & 36.56 & 36.99 \\
 & msmarco-roberta-base-v2 & 68.35 & 62.61 & 66.67 & 30.10 & 31.98 & 32.62 & 56.65 & 29.77 & 46.14 & 43.94 \\
  \midrule
\multirow{1}{*}{colbert ranker} & colbert-v2 & 69.02 & 66.78 & 72.6 & 33.70 & 35.51 & 45.20 & 67.74 & 33.01 & 41.21 & 45.83 \\
\midrule

monoBERT& BERT (340M)  & 70.50& 67.28 & 70.01& 36.88& 31.75& 41.87& 71.36 &31.44 &44.62& 49.35 \\
\midrule
Cohere Rerank-v2& - & 73.22& 67.08& 81.81 &36.36 & 32.51 & 42.51 & 74.44 & 29.60 &47.59 &50.78 \\
\midrule

Promptagator++ & - & &  &76.2 & 37.0  &38.1 & 43.4  & 73.1 &  & &  \\
\midrule
\multirow{2}{*}{TWOLAR } & TWOLAR-Large & 72.82  & 67.61& 84.30&35.70  &33.4 & 47.8 & 75.6& 33.9&52.7  &58.3  \\
 & TWOLAR-XL &73.51  &  70.84   &82.70 & 36.60 &37.1 & 48.0 &76.5 & 33.8&50.8  & 57.9 \\
\midrule
\multirow{2}{*}{RankLLaMA} & LLaMA-2-7B & 74.3 & 72.1 & 85.2 & 30.3 & 40.1 & 48.3 & 73.2 & - & - & - \\
 & LLaMA-2-13B & - & - & 86.1 & 28.4 & 40.6 & 48.7 & 73.0 & - & - & - \\
\bottomrule
\end{tabular} }
\caption{Performance comparison (nDCG@10) of various pointwise reranking models across standard TREC Deep Learning (DL19, DL20) and multiple BEIR benchmark datasets.}
\label{tab:pointwise_app}
\end{table*}

\paragraph{Listwise Reranking Challenges}
\label{sec:listwise}

Listwise methods process queries and documents together, outputting reordered lists (Figure~\ref{fig:Listwiseapproach}) with $O(n)$ complexity. Long inputs exceed LLM context limits, requiring sliding windows or tournament sorts ~\cite{sun2023chatgpt,tamber2023scaling,yoon2024listt5}. Prediction failures, such as missing documents or inconsistent rankings due to input order sensitivity, and reliance on generation APIs ~\cite{ma2023zero,pradeep2023rankvicuna} reduce reliability of listwise reranking.

Table~\ref{tbl:comparison} summarizes the above-mentioned challenges, highlighting sensitivity to input order across methods.

\section{Results and Discussion}
This section evaluates pointwise, pairwise, and listwise reranking methods across IR benchmarks and open-domain datasets, assessing performance, robustness, and generalization to novel queries. It includes three parts: Experimental Setup, Datasets, and Performance Analysis.

\subsection{Experimental Setup}

We compared reranking methods in three categories: pointwise (e.g., MonoT5, RankT5, InRanker, FlashRank), pairwise (e.g., PRP, EcoRank), and listwise (e.g., ListT5, RankGPT, RankVicuna). Pointwise methods score query-document pairs independently, pairwise methods compare document pairs, and listwise methods optimize entire document lists. Models were sourced from public repositories (e.g., Hugging Face) with default settings using Rankify Framework~\cite{abdallah2025rankify} and we integrated the results with RankArena Leaderboard\cite{abdallah2025rankarena}. For the initial retrieval, we used BM25 to pull the top 100 documents per query, which the rerankers then reordered using Pyserini~\cite{Lin_etal_SIGIR2021_Pyserini}. We measured performance with nDCG@10 for TREC DL19, DL20, and BEIR datasets (Covid, NFCorpus, Touche, DBPedia, SciFact, Signal, News, Robust04), and Top-1, Top-10, and Top-50 accuracy for open-domain datasets (Natural Questions and WebQuestions). All experiments were ran on a cluster with NVIDIA A100 GPUs, and we averaged results over three runs with different random seeds to ensure consistency.
\subsection{ Datasets}
We tested rerankers on diverse datasets. TREC DL19 (43 queries) and DL20 (54 queries) simulate web search with graded relevance (0–3). BEIR includes eight datasets—Covid, NFCorpus, Touche, DBPedia, SciFact, Signal, News, Robust04—for zero-shot generalization across domains. Natural Questions (NQ) and WebQuestions (WebQ) test factual retrieval in open-domain QA. FutureQueryEval, with queries unseen until May 2025, evaluates novel query generalization (Section~\ref{sec:FutureQueryEval}). These datasets assess in-domain, out-of-domain, and novel query performance.

\subsection{Performance Analysis}

We analyze pointwise, listwise, and pairwise reranking performance based on Tables~\ref{tab:pointwise_app}, \ref{tab:Listwise}, \ref{tab:PairWise}, and \ref{tab:opendomain}, focusing on key trends and implications for IR systems.

\paragraph{Pointwise Reranking:} Table~\ref{tab:pointwise_app} highlights pointwise methods, which score documents independently. InRanker-3b excels (72.71 on DL19, 81.75 on Covid, and 62.47 on Robust04), leveraging distillation for strong semantic understanding, especially in scientific datasets (78.31 SciFact). MonoT5-3B (71.83 DL19, 80.71 Covid) and TWOLAR-XL (73.51 DL19, 82.70 Covid) follow closely, benefiting from IR-specific fine-tuning. Lighter models like FlashRank-MiniLM (70.40 DL19) and Transformer Ranker-mxbai-rerank-base (84.00 Covid) offer efficiency with competitive accuracy. UPR-T0-3B lags (60.18 DL19), showing zero-shot limitations. All methods struggle on Touche (e.g., InRanker-3b: 29.24) and Signal, likely due to mismatched training data.

\begin{table*}[!t]
\centering
\small
\setlength\tabcolsep{2pt}
\resizebox{0.8\linewidth}{!}{%

\begin{tabular}{l l cc|cccccccc}
\toprule
\textbf{Method} & \textbf{Model} & \textbf{DL19} & \textbf{DL20} & \textbf{Covid} & \textbf{NFCorpus} & \textbf{Touche} & \textbf{DBPedia} & \textbf{SciFact} & \textbf{Signal} & \textbf{News} & \textbf{Robust04} \\

\midrule

InContext & Mistral-7B & 59.2 & 53.6 & 63.9 &33.2  & -& 31.4 &  72.4 &  & &  \\
& Llama-3.18B & 55.7  & 51.9 & 72.8 & 34.7 & -& 35.3 & 76.1 &  & &  \\
\midrule
\multirow{5}{*}{RankGPT} & Llama-3.2-1B & 47.13 & 44.93 & 59.29 & 31.43 & 37.20 & 26.04 & 67.49 & 26.94 & 38.76 & 38.14 \\
 & Llama-3.2-3B & 58.05 & 53.33 & 68.18 & 31.86 & 37.23 & 36.14 & 67.32 & 32.75 & 42.49 & 44.83 \\

& gpt-3.5-turbo&  65.80 &62.91 &76.67& 35.62 &36.18 &44.47& 70.43 &32.12& 48.85& 50.62\\
&gpt-4 & 75.59& 70.56 &85.51 &38.47& 38.57 &47.12& 74.95& 34.40 &52.89 &57.55\\

&llama 3.1 8b &58.46 & 59.68 & 69.61 & 33.62 & 37.98  & 37.25 & 69.82 & 32.95 &43.90 &49.59 \\

\midrule
\multirow{2}{*}{ListT5} & listt5-base & 71.80  & 68.10 & 78.30 & 35.60 &  33.40 & 43.70 & 74.10 & 33.50 & 48.50 & 52.10\\
 & listt5-3b & 71.80 & 69.10 & 84.70 & 37.70 & 33.60 & 46.20 & 77.0 &  33.80 & 53.20 & 57.80 \\ 
  \midrule
\multirow{6}{*}{lit5dist} & LiT5-Distill-base & 72.46 & 67.91 & 70.48 & 32.60 & 33.69 & 42.78 & 56.35 & 34.16 & 41.53 & 44.32 \\
 & LiT5-Distill-large & 73.18 & 70.32 & 73.71 & 34.95 & 33.46 & 43.17 & 66.70 & 30.88 & 44.41 & 52.46 \\
 & LiT5-Distill-xl & 72.45 & 72.46 & 72.97 & 35.81 & 32.76 & 43.52 & 71.88 & 31.23 & 46.59 & 53.77 \\
 & LiT5-Distill-base-v2 & 71.63 & 69.13 & 70.53 & 34.23 & 34.25 & 43.18 & 67.24 & 33.28 & 45.25 & 48.00 \\
 & LiT5-Distill-large-v2 & 72.15 & 67.78 & 73.10 & 34.05 & 34.55 & 43.35 & 69.30 & 31.16 & 42.42 & 50.95 \\
 & LiT5-Distill-xl-v2 & 71.94 & 71.93 & 73.08 & 34.68 & 34.29 & 44.59 & 69.68 & 32.76 & 45.88 & 51.70 \\
  \midrule
\multirow{3}{*}{lit5score}& LiT5-Score-base & 68.59 & 66.04 & 66.47 & 32.72 & 32.84 & 36.49 & 57.52 & 24.01 & 41.44 & 45.12 \\
 & LiT5-Score-large & 71.01 & 66.43 & 69.84 & 33.64 & 30.71 & 37.85 & 62.48 & 24.81 & 43.35 & 47.42 \\
 & LiT5-Score-xl & 69.36 & 65.56 & 69.66 & 34.36 & 29.09 & 39.10 & 67.50 & 24.07 & 44.95 & 52.88 \\
  \midrule
\multirow{1}{*}{Vicuna Reranker} & Vicuna 7b  & 67.19 & 65.29 & 78.30 & 32.95 & 32.71 & 43.28 & 70.49 & 32.87 & 44.98 & 47.83 \\
 \midrule
\multirow{1}{*}{Zephyr Reranker} & Zephyr 7B & 74.22 & 70.21 & 80.70 & 36.58 & 31.12 & 43.18 & 75.13 & 31.96 & 48.95 & 54.20 \\
 \midrule
 \multirow{4}{*}{Setwise} & Flan-T5-Large (heapsort) & 67.0 & 61.8 & 76.8 & 32.5 & 30.3 & 41.3 & 62.0 & 31.9 & 43.9 & 46.2 \\
 & Flan-T5-XL (heapsort) & 69.3 & 67.8 & 75.7 & 35.2 & 28.3 & 42.8 & 67.7 & 31.4 & 46.5 & 52.0 \\
 & Flan-T5-Large (bubblesort) & 67.8 & 62.4 & 76.1 & 33.8 & 39.4 & 44.1 & 63.6 & 35.1 & 44.7 & 49.7 \\
 & Flan-T5-XXL (bubblesort) & 71.1 & 68.6 & 76.8 & 34.6 & 38.8 & 42.4 & 75.4 & 34.3 & 47.9 & 53.4 \\
\midrule
\multirow{3}{*}{TourRank} & GPT-3.5-turbo (TourRank-1) & 66.23 & 63.74 & 77.17 & 36.35 & 29.38 & 40.62 & 69.27 & 29.79 & 46.41 & 52.70 \\
 & GPT-3.5-turbo (TourRank-2) & 69.54 & 65.20 & 79.85 & 36.95 & 30.58 & 41.95 & 71.91 & 31.02 & 48.13 & 55.27 \\
 & GPT-3.5-turbo (TourRank-10) & 71.63 & 69.56 & 82.59 & 37.99 & 29.98 & 44.64 & 72.17 & 30.83 & 51.46 & 57.87 \\
\midrule

\end{tabular} }
\caption{Evaluation results (nDCG@10) of listwise reranking approaches on TREC Deep Learning (DL19, DL20) and selected BEIR benchmarks. For the full model list and comparison, please refer Appendix~\ref{apndeix:ListWise}}
\label{tab:Listwise}
\end{table*}

\begin{table}[!t]
\centering
\small
\setlength\tabcolsep{2pt}
\resizebox{\linewidth}{!}{%

\begin{tabular}{l l cc|cccccccc}
\toprule
\textbf{Method} & \textbf{Model} & \textbf{DL19} & \textbf{DL20} & \textbf{Covid} & \textbf{NFCorpus} & \textbf{Touche} & \textbf{DBPedia} & \textbf{SciFact} & \textbf{Signal} & \textbf{News} & \textbf{Robust04} \\
\midrule
\multirow{3}{*}{PRP} & FLAN-T5-XL &68.66 &66.59 & 77.58 &- & 40.48 & 44.77 & 73.43  &35.62 & 46.45& 50.74 \\ 
 & FLAN-T5-XXL&67.00 &67.35 & 74.39 &-  &41.60 & 42.19 & 72.46 & 35.12 & 47.26 &52.38 \\
  & FLAN-UL2& 72.65 & 70.46 & 79.45&-& 37.89 &46.47 &73.33& 35.20& 49.11& 53.43  \\
\midrule

EchoRank & Flan-T5-Large &60.29 & 58.24 & 54.98 & 30.06 & 24.93& 35.18   & 65.09 & 33.86 & 19.26 & 21.51 \\
& Flan-T5-xl &59.62 & 58.98 & 55.51 & 30.57&25.24 &  35.49 & 69.14 &  33.11& 19.05&  21.62\\
\bottomrule

\end{tabular} }
\caption{Performance outcomes (nDCG@10) of pairwise reranking methods evaluated on TREC DL benchmarks and various BEIR datasets. }
\label{tab:PairWise}
\end{table}
\paragraph{Listwise Reranking:} Table~\ref{tab:Listwise} shows listwise methods, which optimize document interactions. RankGPT-gpt-4 leads (75.59 DL19, 85.51 Covid), excelling in nuanced relevance. Zephyr-7B (74.22 DL19, 80.70 Covid) and ListT5-3b (71.80 DL19, 84.70 Covid) perform strongly, balancing accuracy and efficiency. LiT5-Distill-xl (72.45 DL19) scales well, but smaller models like InContext-Mistral-7B (59.2 DL19) falter due to context constraints. Touche remains challenging (e.g., ListT5-3b: 33.60), reflecting difficulties with argumentative queries.

\paragraph{Pairwise Reranking:} Table~\ref{tab:PairWise} evaluates pairwise methods, comparing document pairs. PRP-FLAN-UL2 performs best (72.65 DL19, 79.45 Covid), adept at fine-grained judgments but less scalable due to quadratic complexity. EcoRank-Flan-T5-xl (59.62 DL19, 55.51 Covid) prioritizes efficiency, sacrificing accuracy. Pairwise methods underperform on Touche (e.g., PRP-FLAN-UL2: 37.89), struggling with subjective relevance.

For additional comparisons, including smaller model variants and complete tables, see Appendix~\ref{apndeix:pointwise} and Appendix~\ref{apndeix:ListWise} (Tables~\ref{tab:pointwise_app} and~\ref{tab:Listwise_app}).

\begin{table}[!ht]

\centering
\resizebox{0.45\textwidth}{!}{
\begin{tabular}{@{}l |l| c c c  | c c c@{}}\toprule
Reranking/ &  Model  &  \multicolumn{3}{c}{NQ} & \multicolumn{3}{c}{WebQ}  \\

& & Top-1 & Top-10 & Top-50 & Top-1 & Top-10 & Top-50   \\
\midrule
BM25      & - & 23.46 & 56.32 & 74.57 & 19.54  & 53.44 & 72.34 \\

\midrule
 
\multirow{2}{*}{UPR}  

  & T0-3B & 35.42 & 67.56 & 76.75 & 32.48 & 64.17 & 73.67 \\

  &gpt-neo-2.7B &28.75  & 64.81  &  76.56 &24.75    & 59.64 &  72.63  \\

\midrule

 \multirow{1}{*}{RankGPT} & llamav3.1-8b & 41.55 & 66.17 & 75.42 & 38.77 & 62.69 & 73.12 \\
 
\midrule

 \multirow{4}{*}{FlashRank} 
  & TinyBERT-L-2-v2& 31.49 & 61.57 & 74.95 & 28.54 & 60.62 & 73.17 \\
  & MultiBERT-L-12 & 11.99&  43.54 & 69.63 & 12.54  & 45.91 & 67.91 \\
  & ce-esci-MiniLM-L12-v2 & 34.70 & 64.81 & 76.17 & 31.84 & 62.54 & 73.47  \\
 & T5-flan & 7.95  & 36.14  & 66.67 &   12.05 & 42.96   & 67.27  \\
\midrule
  
  \multirow{3}{*}{RankT5}
  & base & 43.04 & 68.47 & 76.28 & 36.95 & 64.27 & 74.45  \\
  & large &45.54 & 70.02 & 76.81 & 38.77 & 66.48 & 74.31 \\
  &3b & 47.17& 70.85 & 76.89 & 40.40 & 66.58 & 74.45  \\

\midrule

\multirow{3}{*}{Inranker}  
& small & 15.90 & 46.84 & 69.83  & 14.46  & 46.25 & 69.98 \\
&base  &15.90 & 48.11 & 69.66 &14.46 & 46.80 & 69.68 \\
&3b &15.90 & 48.06 & 69.00 & 14.46 & 46.11 & 69.34 \\
\midrule

\multirow{1}{*}{LLM2Vec} 
&Meta-Llama-31-8B & 24.32 & 59.55 & 75.26  &  26.72 &60.48 & 73.47\\

\midrule

\multirow{1}{*}{MonoBert} 
& large & 39.05 & 67.89 & 76.56 & 34.99 &64.56 & 73.96 \\

\midrule

\multirow{1}{*}{Twolar} 
& twolar-xl & 46.84 & 70.22 & 76.86 &41.68 &67.07 &74.40 \\

\midrule

\multirow{2}{*}{Echorank} 
& flan-t5-large& 36.73 & 59.11 & 62.38 & 31.74 &58.75 &61.51 \\
&flan-t5-xl&   41.68 & 59.05 & 62.38  & 36.22 &57.18 &61.51\\

\midrule

\multirow{1}{*}{\makecell[l]{Incontext Reranker}} &  \multirow{1}{*}{llamav3.1-8b} & \multirow{1}{*}{15.15} & \multirow{1}{*}{57.11}  & \multirow{1}{*}{76.48}  & \multirow{1}{*}{18.89}   & \multirow{1}{*}{52.16} &  \multirow{1}{*}{71.70} \\

\midrule

\multirow{4}{*}{Lit5}
&  LiT5-Distill-base  & 40.05 & 65.95 & 75.73  & 36.76 & 63.48 & 73.12 \\
&  LiT5-Distill-large  & 44.40 & 67.59 & 76.01   &  39.66  &64.56 &73.67\\
&  LiT5-Distill-large-v2  & 46.53 &  67.83 & 75.87  & 41.97 &65.64 &72.98 \\
&  LiT5-Distill-xl-v2  & 47.92 & 69.03 & 76.17  & 41.53 &65.69 &73.27  \\

\midrule

\multirow{4}{*}{ \makecell[l]{Sentence Transformer} }
& GTR-large & 40.63 &68.25 & 76.73  & 38.97 &65.30 &73.57 \\
& T5-large & 30.80 &63.35 &76.37   & 30.51  &61.71 &73.37 \\
& GTR-xxl   & 42.93 &  68.55&  77.00 &39.41    & 65.89 & 74.01 \\
&  T5-xxl  & 38.89  &  67.78  &76.64   &  35.82  & 65.20  & 74.01 \\
\midrule

\end{tabular}}
\caption{Performance of re-ranking methods on BM25-retrieved documents for NQ Test and WebQ Test. Results are reported in terms of Top-1, Top-5, Top-10, Top-20, and Top-50 accuracy. Please note that some results may differ from the original papers (e.g., UPR) as our experiments were conducted with the top 100 retrieved documents, whereas the original studies used 1,000 documents for ranking.
}
\label{tab:opendomain}
\end{table}

\paragraph{Reranking on Open-Domain QA:} Table~\ref{tab:opendomain} evaluates reranking on open-domain QA tasks (NQ, WebQ). BM25 baselines at 23.46\% Top-1 (NQ) and 19.54\% (WebQ). Fine-tuned models excel: LiT5-Distill-xl-v2 (47.92\% NQ, 41.53\% WebQ), RankT5-3b (47.17\% NQ, 40.40\% WebQ), and TWOLAR-XL (46.84\% NQ, 41.68\% WebQ) lead, adept at factual queries. RankGPT-llamav3.1-8b (41.55\% NQ) follows. Efficient models like FlashRank-MiniLM (34.70\% NQ, 31.84\% WebQ) balance performance and speed. EcoRank-Flan-T5-XL (41.68\% NQ) and InContext-llamav3.1-8b (15.15\% NQ) lag, struggling with QA contexts. Pointwise and listwise methods outperform pairwise in open-domain tasks.

For additional comparisons, including smaller model variants and complete tables, see Appendix~\ref{apndeix:open-domain} (Tables~\ref{tab:opendomain_app}).

\subsection{Discussion and Implications}

Our comprehensive evaluation reveals several novel insights beyond traditional performance reporting:

\textbf{Temporal Generalization Gap:} Comparing performance on established benchmarks (Tables 2-4) versus FutureQueryEval (Tables 6-8) reveals a consistent 5-15\% performance drop across all method categories, indicating significant temporal sensitivity in reranking models. This suggests that claims of "generalization" based on standard benchmarks may be overstated.

\textbf{Method-Specific Degradation Patterns:} Listwise methods show the smallest performance drop on novel queries (avg. 8\% degradation) compared to pointwise (12\%) and pairwise (15\%) methods, suggesting that inter-document modeling provides better robustness to unseen content.

\textbf{Scale vs. Robustness Trade-off:} While larger models generally perform better on established benchmarks, the performance gap narrows significantly on FutureQueryEval, indicating diminishing returns of scale for novel query generalization.

\textbf{Domain-Specific Vulnerabilities:} All methods struggle disproportionately with argumentative (Touche) and informal (Signal) domains, suggesting systematic gaps in current training paradigms rather than random performance variations.

\section{FutureQueryEval}
\label{sec:FutureQueryEval}

Reranking models are expected to generalize beyond their pretraining corpora, yet most benchmarks—like TREC DL19, DL20, and BEIR—contain queries and documents that may overlap with the training data of LLMs~\citep{yu2022generate}. To ensure evaluations are performed on unseen, truly novel content, we introduce \textbf{FutureQueryEval}, a dataset designed to test reranking models on queries and documents collected after April 2025. The dataset comprises 148 queries across seven diverse topical categories (e.g., Technology, Sports, Politics). Each query is paired with manually annotated documents collected post-April 2025, ensuring they are out-of-distribution for most existing LLMs. Relevance was assigned using a 3-level scale: 0 (irrelevant), 1 (partial), 2 (highly relevant). We validated novelty against GPT-4, confirming queries refer to events beyond the model's knowledge. Full construction details, corpus examples, and statistical breakdowns (e.g., document lengths, token CDF, query distributions) are provided in Appendix~\ref{sec:dataset_details}.

\section{Results on FutureQueryEval}
\label{sec:futureeval-results}
This section digs into how well our reranking models perform on the FutureQueryEval dataset, a new benchmark we introduced to test generalization on queries and documents unseen by models until May 2025. We evaluate pointwise, listwise, and pairwise reranking methods, focusing on their ability to handle novel content without the risk of training data contamination. Using metrics like NDCG@10 from Tables~\ref{tab:pointwise_our}, \ref{tab:listwise_our}, and \ref{tab:pairwise_our}, and MAP scores from Figure~\ref{fig:map1_pointwise}, ~\ref{fig:map2_pointwise} and ~\ref{fig:map3_listwise}, we highlight key trends, compare approaches, and discuss what the results mean for building robust IR systems. We provide detailed NDGC, MAP trends provided in Appendix~\ref{sec:appendix-fqe-results}

\paragraph{Pointwise:}
MonoT5-3B-10k achieves the best performance (NDCG@10: 60.75), followed closely by Twolar-xl (60.03), confirming the advantage of large-scale fine-tuning. Efficient models like FlashRank (55.43) and ColBERT-v2 (54.20) offer strong trade-offs between speed and accuracy. UPR and GPT2 variants underperform (e.g., T5-small at 47.24), likely due to limited or zero-shot tuning.

\paragraph{Listwise:} Listwise models benefit from inter-document reasoning. Zephyr-7B leads with NDCG@10 of 62.65, while Vicuna-7B (58.63) also performs well. ListT5 models fail to generalize (e.g., ListT5-3B: 9.72), possibly due to misalignment between their training data and the current query domain. RankGPT and InContext rerankers show moderate performance but lag behind the top models.

\paragraph{Pairwise.} EchoRank, powered by Flan-T5, demonstrates promising results (NDCG@10: 54.97 for XL), nearly matching pointwise methods while offering stronger pairwise relevance signals. However, the computational cost of pairwise comparisons limits scalability.

\begin{table}[]
    \centering
\setlength\tabcolsep{2pt}
\resizebox{.95\linewidth}{!}{%
\begin{tabular}{ll|rrrrrr} 
\toprule
Method & Model & NDCG@1 & NDCG@5 & NDCG@10 & NDCG@20 & NDCG@50 & NDCG@100 \\
\midrule
BM25 & - & 39.86 & 43.01 & 46.42 & 50.26 & 53.79 & 55.44 \\

\midrule
\multirow{1}{*}{upr} 
 & t0-3b & 44.93 & 47.98 & 52.16 & 56.82 & 59.43 & 59.58 \\

  \midrule
\multirow{1}{*}{flashrank} 
 & MiniLM-L-12-v2 & 53.04 & 51.72 & 55.43 & 59.64 & 61.95 & 62.21 \\
  \midrule
\multirow{1}{*}{monot5} & base & 53.04 & 55.15 & 57.88 & 61.61 & 63.58 & 63.7 \\
  
  \midrule
\multirow{1}{*}{inranker} 
 & inranker-3b & 18.24 & 28.47 & 32.39 & 38.11 & 41.22 & 44.5 \\
  \midrule
\multirow{1}{*}{transformer ranker} 
 & MiniLM-L-6-v2 & 53.04 & 51.94 & 55.51 & 59.55 & 61.95 & 62.23 \\
 
  \midrule
\multirow{1}{*}{Splade reranker} & splade-cocondenser & 46.96 & 49.02 & 52.93 & 57.72 & 60.08 & 60.21 \\
 \midrule
\multirow{1}{*}{\makecell[l]{Sentence Transformer }} 
 & gtr-t5-large & 45.95 & 49.43 & 52.97 & 57.7 & 60.08 & 60.27 \\
  \midrule
\multirow{1}{*}{colbert ranker} & colbert-v2 & 50.34 & 50.54 & 54.2 & 58.65 & 61.05 & 61.4 \\
  \midrule
\multirow{1}{*}{monobert} & monobert-large & 50.0 & 53.39 & 56.99 & 61.16 & 63.03 & 63.11 \\
 \midrule
\multirow{1}{*}{llm2vec} 
 & Mistral-7B-Instruct-v0.2  & 45.61 & 48.69 & 53.64 & 58.19 & 60.4 & 60.56 \\
  \midrule
\multirow{1}{*}{twolar} & twolar-xl & 55.41 & 57.79 & 60.03 & 63.68 & 65.19 & 65.23 \\

\midrule
\multirow{2}{*}{RankLLaMA} & LLaMA2-7B & 54.81 & 57.53 & 61.09 & 63.16 & 63.47 \\
 & LLaMA2-13B & 55.72 & 59.00 & 61.94 & 63.66 & 63.99 \\

\bottomrule
\end{tabular} }

    \caption{ Performance of pointwise reranking methods on the FutureQueryEval dataset. Metrics reported include NDCG  at different cutoffs (1, 5, 10, 20, 50, and 100).}
    \label{tab:pointwise_our}
\end{table}

\subsection{Overall Findings and Implications}

FutureQueryEval reveals clear trends in how reranking methods handle unseen, novel content. Listwise models such as Zephyr-7B (NDCG@10: 62.65) and Vicuna-7B (58.63) lead performance by modeling document interactions, making them ideal for complex, context-rich queries. Among pointwise models, MonoT5-3B (60.75) and Twolar-xl (60.03) offer strong generalization and efficiency, especially when fine-tuned on large IR datasets. EchoRank's pairwise method also improves notably (Flan-T5-XL: 54.97), though its higher computational cost may limit scalability. The newly evaluated LLM-based methods demonstrate strong generalization capabilities on FutureQueryEval. TourRank with GPT-4o achieves the highest performance among listwise methods (NDCG@10: 62.02), outperforming many established approaches and confirming the superior generalization of advanced LLMs. RankLLaMA-13B shows competitive pointwise performance (NDCG@10: 59.00), while Setwise methods provide a balanced approach with Flan-T5-Large achieving 56.57 NDCG@10 through heapsort aggregation. These results suggest that listwise rerankers are best suited for high-accuracy scenarios (e.g., news, healthcare), while pointwise models like FlashRank provide reliable performance with lower resource demands. Pairwise approaches, if optimized, can bridge precision and robustness. Poor performance by some models (e.g., ListT5: 9.72, UPR T5-small: 47.24) highlights a key challenge—many rerankers rely on training data misaligned with emerging topics. However, advanced LLM-based methods like TourRank with GPT-4o (62.02) and RankLLaMA-13B (59.00) demonstrate superior generalization to novel content, supporting the hypothesis that larger, more capable LLMs exhibit better zero-shot transfer to unseen queries. FutureQueryEval thus underscores the need for evolving benchmarks and hybrid reranking strategies that blend pointwise speed, listwise reasoning, and pairwise precision.

\begin{table}[]
    \centering
\setlength\tabcolsep{2pt}
\resizebox{0.95\linewidth}{!}{%
\begin{tabular}{ll|rrrrrr}
\toprule
Method & Model & NDCG@1 & NDCG@5 & NDCG@10 & NDCG@20 & NDCG@50 & NDCG@100 \\
\midrule
\multirow{1}{*}{listt5} & listt5-base & 12.84 & 13.28 & 11.5 & 11.12 & 25.39 & 30.83 \\
  \midrule
\multirow{1}{*}{incontext reranker} & Mistral-7B-Instruct-v0.2 & 14.86 & 21.46 & 22.92 & 27.46 & 35.9 & 40.04 \\
 \midrule
\multirow{1}{*}{vicuna reranker} 
 & rank vicuna 7b v1 noda & 56.08 & 55.12 & 58.63 & 62.68 & 64.24 & 64.6 \\
  \midrule
\multirow{1}{*}{zephyr reranker} & rank zephyr 7b v1 full & 59.46 & 60.5 & 62.65 & 65.57 & 67.0 & 67.14 \\
\midrule
\multirow{1}{*}{rankgpt} 
 & llamav3.1-8b & 39.86 & 52.19 & 54.18 & 57.71 & 60.15 & 60.66 \\
 \midrule
\multirow{1}{*}{llm layerwise ranker} 
 & bge-reranker-v2.5-gemma2 & 32.77 & 32.42 & 34.85 & 39.15 & 44.99 & 47.96 \\
   \midrule
   \multirow{2}{*}{Setwise} & Flan-T5-Large (heapsort) & 53.51 & 56.57 & 59.43 & 61.94 & 62.34 \\
 & Flan-T5-Large (bubblesort) & 53.45 & 55.84 & 58.82 & 61.46 & 62.08 \\
\midrule
\multirow{2}{*}{TourRank} & LLaMA3-8B (TourRank-10) & 54.96 & 57.06 & 57.93 & 57.93 & 57.93 \\
 & GPT-4o (TourRank-10) & 59.06 & 62.02 & 63.53 & 65.73 & 65.86 \\
 \bottomrule    
\end{tabular} }

\caption{Performance of listwise, setwise, and tournament-based reranking methods evaluated on the FutureQueryEval dataset.}
    \label{tab:listwise_our}
\end{table}

\begin{table}[]
    \centering
\resizebox{0.95\linewidth}{!}{%
\begin{tabular}{ll|rrrrrr} 
\toprule
Method & Model & NDCG@1 & NDCG@5 & NDCG@10 & NDCG@20 & NDCG@50 & NDCG@100 \\
 \midrule
\multirow{2}{*}{Echorank} & flan-t5-large & 54.05 & 54.25 & 54.8 & 57.08 & 57.41 & 57.41 \\
 & flan-t5-xl & 57.09 &  54.45 & 54.97 & 57.41 & 57.77 &  57.77 \\
\midrule
\end{tabular} }

    \caption{Evaluation of pairwise reranking methods using FutureQueryEval. }
    \label{tab:pairwise_our}
\end{table}

\subsection{Efficiency-Effectiveness Trade-Off}

\begin{figure}[h]
    \centering
    \includegraphics[width=0.5\textwidth]{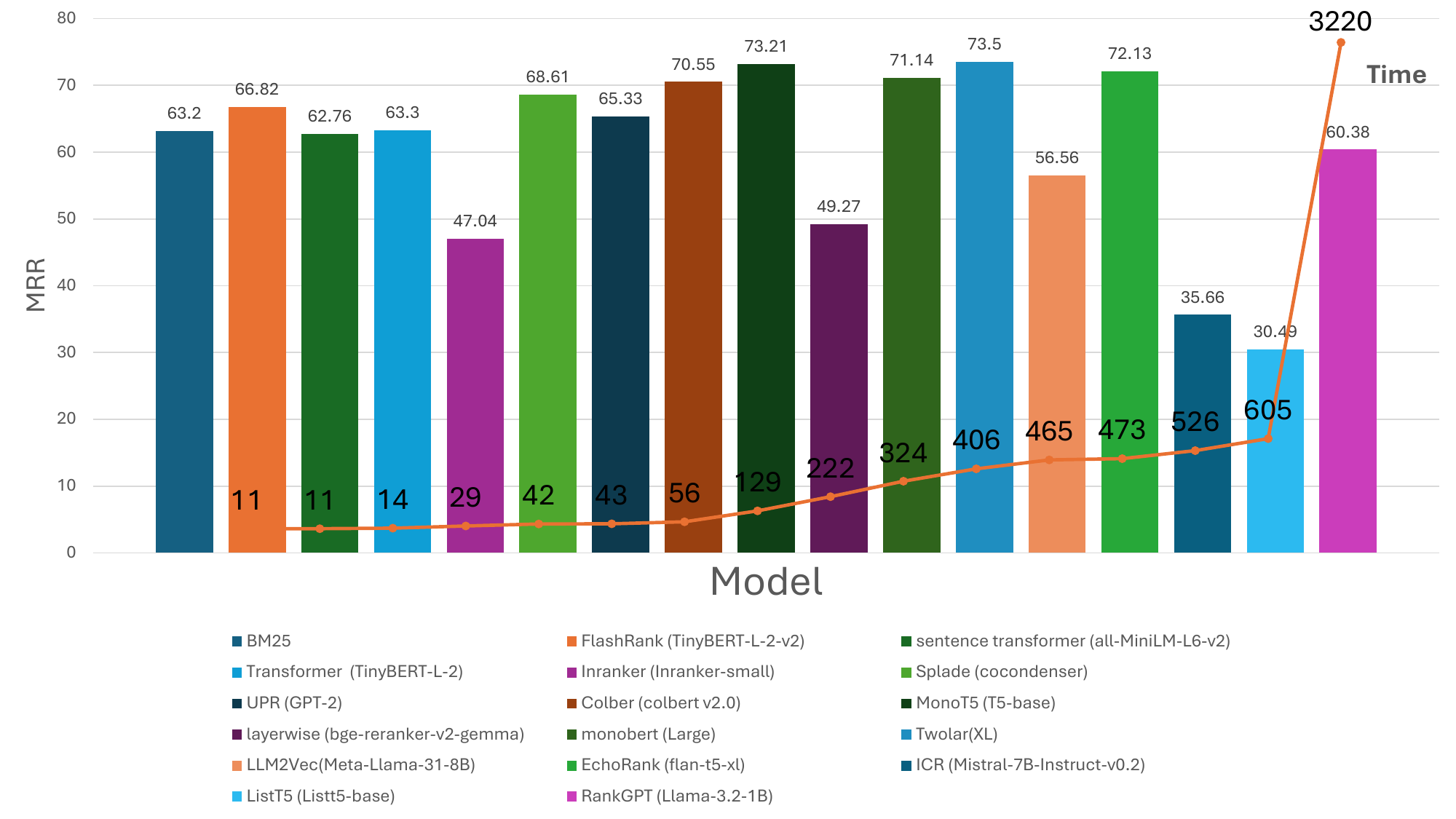}
    \caption{MRR vs. Time for the fastest model from each reranking method on FutureQueryEval, highlighting efficiency-effectiveness trade-offs.}
    \label{fig:mrr_vs_time_best_time}
\end{figure}

\begin{figure}[h]
    \centering
    \includegraphics[width=0.5\textwidth]{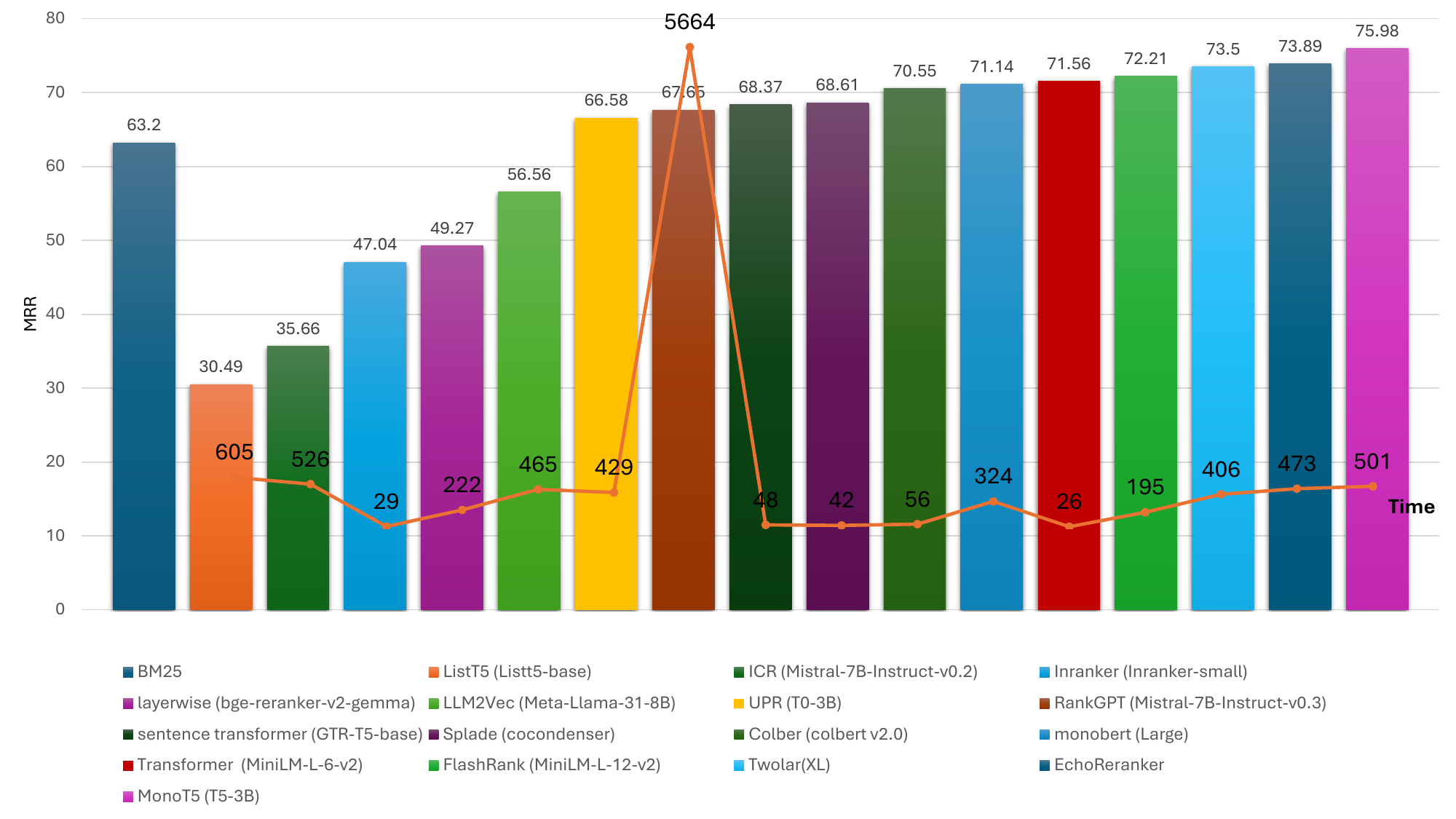}
    \caption{MRR vs. Time for the highest-MRR model from each reranking method on FutureQueryEval, showcasing peak performance and associated time costs.}
    \label{fig:best_mrr_vs_time}
\end{figure}

Beyond ranking quality, practical IR systems must balance effectiveness and runtime. We compare reranking models on FutureQueryEval using Mean Reciprocal Rank (MRR) and processing time. Figures~\ref{fig:mrr_vs_time_best_time} and~\ref{fig:best_mrr_vs_time} illustrate the trade-off between speed and accuracy by selecting the fastest and highest-MRR models per method. Figure~\ref{fig:mrr_vs_time_best_time} shows that Sentence Transformer (all-MiniLM-L6-v2) is the fastest (11.72s, MRR: 62.76), while FlashRank (TinyBERT) offers better accuracy (66.82) at similar speed. Transformer Ranker (TinyBERT) is also efficient (14.23s, 63.30 MRR). In contrast, RankGPT (Llama-3.2-1B) takes over 53 minutes for a modest 60.38 MRR. MonoT5-base provides a good balance (129.91s, 73.21 MRR). Figure~\ref{fig:best_mrr_vs_time} highlights top-performing models. MonoT5-3B (75.98 MRR) and Twolar-xl (73.50) are most effective but slower. FlashRank (MiniLM) again shows a strong middle ground (72.21 MRR, 195.48s). Transformer Ranker (MiniLM-L6-v2) is the fastest among high performers (26.65s, 71.56 MRR). RankGPT and ListT5 lag in both metrics, demonstrating inefficiency.

\section{Conclusion}

We presented a comprehensive empirical study of 22 reranking methods across 40 variants, spanning pointwise, pairwise, and listwise paradigms. Our evaluation across standard IR benchmarks and the novel FutureQueryEval dataset reveals that while LLM-based rerankers excel on familiar queries, their generalization to unseen queries remains inconsistent. Lightweight models, particularly those fine-tuned on IR data, achieve strong trade-offs between accuracy and efficiency. FutureQueryEval, a temporally novel benchmark, exposes critical limitations in current reranking methods when faced with truly unseen data. Listwise approaches, especially Zephyr-7B and Vicuna-7B, achieve the highest effectiveness but at significant computational cost. Pointwise rerankers like MonoT5-3B and Twolar-XL offer scalable, high-performing alternatives. Pairwise methods provide fine-grained relevance signals yet struggle to scale.

\section{Limitations}

Despite the promising performance of LLM-based rerankers, several limitations remain. First, the computational overhead of prompting and decoding with large language models like GPT-4 can be significant, particularly during inference when reranking large document pools. This hinders real-time applicability and increases the environmental cost of deployment.

Second, LLMs are prone to hallucination and may generate plausible but incorrect rationales when producing pairwise or listwise justifications. This challenges the trustworthiness of model explanations in high-stakes applications such as legal or medical document retrieval.

Third, current LLM reranking approaches often rely on zero-shot or few-shot prompting strategies that do not generalize well to highly domain-specific or low-resource datasets. The lack of fine-grained control over ranking behavior makes it difficult to enforce consistency or incorporate explicit user preferences. 

Finally, many LLM-based approaches assume access to powerful proprietary APIs (e.g., OpenAI’s GPT-4), which raises concerns about reproducibility, data privacy, and fairness in academic and industrial settings where such access may not be uniformly available.

\bibliography{custom}

\appendix
\section{FutureQueryEval Dataset}
\subsection{Dataset Details}
\label{sec:dataset_details}

This section provides a comprehensive overview of how the \textbf{FutureQueryEval} dataset was built and analyzed.

As a pioneering step, we introduce \textbf{FutureQueryEval}, a novel test set comprising 148 queries and their associated documents, collected from April 2025 onward. This dataset is designed to include content published after the training cutoff of most existing LLMs, ensuring that the queries and passages remain unknown to these models until May 2025. To verify this, we tested a subset of queries against \texttt{GPT-4}, confirming their novelty. For example, the query \emph{"What specific actions has Egypt taken to support injured Palestinians from Gaza, as highlighted during the visit of Presidents El-Sisi and Macron to Al-Arish General Hospital?"} relates to events from April 2025, which are inaccessible to LLMs trained on data prior to this period. The dataset spans seven categories: World News \& Politics, Technology, Sports, Science \& Environment, Business \& Finance, Health \& Medicine, and Entertainment \& Culture, reflecting a diverse range of topics.

The corpus was constructed by collecting paragraph-length documents from online sources published after April 2025, similar to the example \emph{"Achieving sustainable development depends on fostering innovation through collective action. Governments must set ambitious frameworks, businesses should invest in green solutions, and young people and start-ups need to drive fresh ideas..."}. For each query, we retrieved an initial set of candidate documents using a general-purpose search engine, followed by manual relevance annotations. The author of the paper who annotated this dataset. Relevance labels were assigned as follows: 0 for irrelevant, 1 for partially relevant, and 2 for highly relevant, based on expert judgment. This process resulted in a qrels file linking queries to documents with their relevance scores, totaling 2,938 query-document pairs across 2,787 unique documents. 

\begin{figure}[h]
    \centering
    \includegraphics[width=0.5\textwidth]{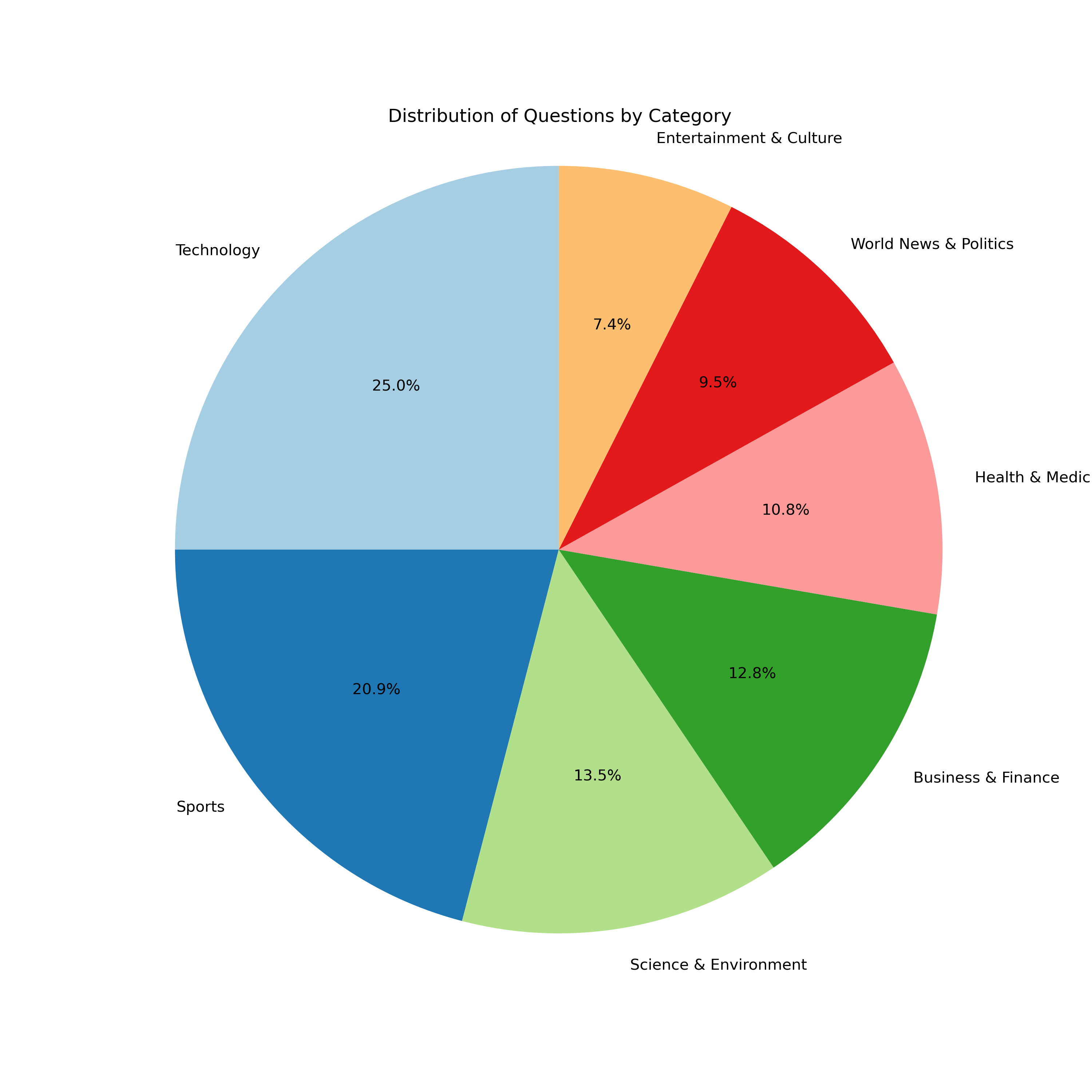}
    \caption{Distribution of Questions by Category in FutureQueryEval}
    \label{fig:category_distribution}
\end{figure}

To provide a comprehensive statistical overview, we conducted several analyses on the dataset. First, we examined the distribution of questions across categories, revealing a balanced yet varied composition. Approximately 9.5\% of queries fall under World News \& Politics, 25.0\% under Technology, 20.9\% under Sports, 13.5\% under Science \& Environment, 12.8\% under Business \& Finance, 10.8\% under Health \& Medicine, and 7.4\% under Entertainment \& Culture. This distribution is visualized in Figure~\ref{fig:category_distribution}, which highlights the diversity of query topics and supports the dataset's applicability across multiple domains.

\begin{figure}[h]
    \centering
    \includegraphics[width=0.5\textwidth]{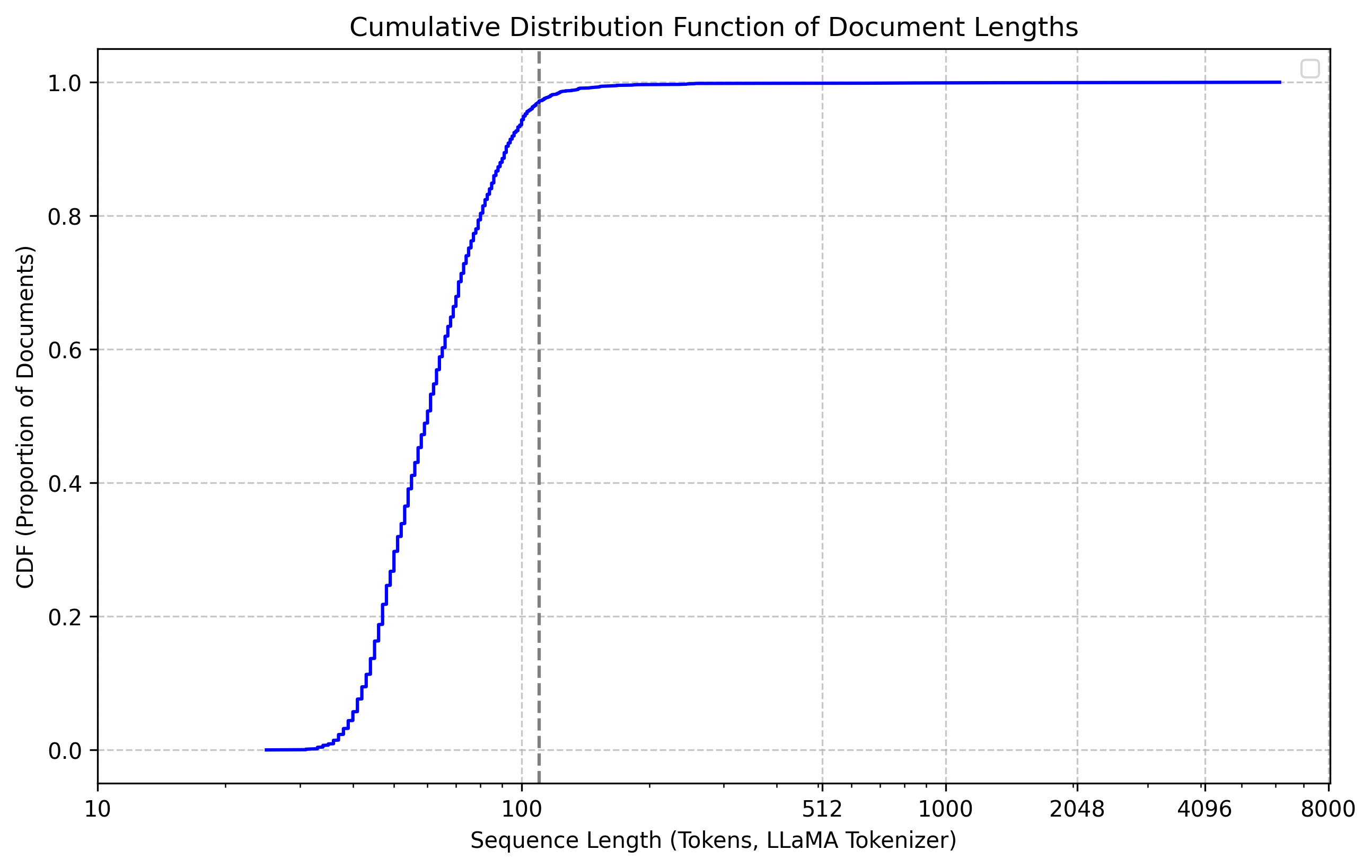}
    \caption{Cumulative Distribution Function of Document Lengths in FutureQueryEval (using LLaMA tokenizer)}
    \label{fig:document_length_cdf}
\end{figure}

Second, we analyzed the length distribution of documents in the corpus using the LLaMA tokenizer to count tokens, ensuring alignment with modern NLP practices. The cumulative distribution function (CDF) of document lengths, shown in Figure~\ref{fig:document_length_cdf}, indicates that 97\% of documents have a length of fewer than 110 tokens, with a maximum length of 6,138 tokens. This analysis ensures that the dataset is compatible with typical IR model input constraints, while also identifying a cutoff for excluding the longest 3\% of documents to maintain clarity, aligning with practices in datasets like MS MARCO~\citep{nguyen2016ms,nogueira2019passage}.

The dataset contains 2,787 unique documents and 2,938 total query-document pairs. Relevance annotations show that each query has, on average, 6.54 relevant passages. To ensure compatibility with transformer models, we analyzed document lengths using the LLaMA tokenizer. Figure~\ref{fig:document_length_cdf} shows that 97\% of documents are shorter than 110 tokens, aligning with modern IR benchmarks like MS MARCO~\citep{nguyen2016ms}.

\begin{table}[h]
    \centering
    \begin{tabular}{lc}
        \toprule
        \textbf{Metric} & \textbf{Value} \\
        \midrule
        Total Queries & 148 \\
        Total Documents & 2,787 \\
        Total Query-Document Pairs & 2,938 \\
        Avg. Relevant Docs per Query & 6.54 \\
        97th Percentile Doc Length (tokens) & 110 \\
        Max Doc Length (tokens) & 6,138 \\
        \bottomrule
    \end{tabular}
    \caption{Summary Statistics of FutureQueryEval}
    \label{tab:dataset_stats}
\end{table}

\subsection{Results}
\label{sec:appendix-fqe-results}

This section provides extended analysis for the results shown in Section~\ref{sec:futureeval-results}, covering performance trends across pointwise, listwise, and pairwise reranking models using NDCG and MAP metrics.

\begin{figure}[htbp]
    \centering
    \includegraphics[width=0.5\textwidth]{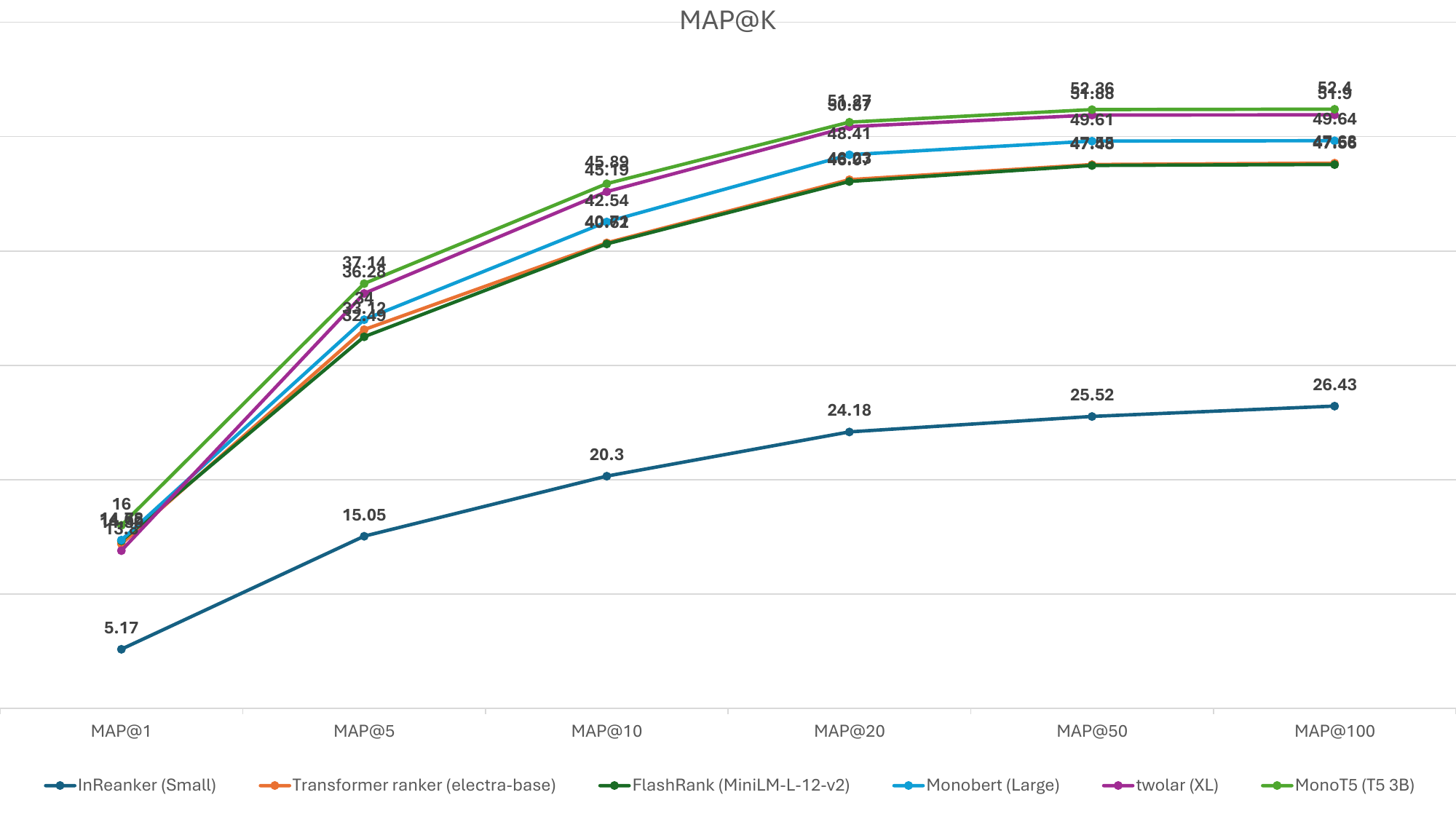}
    \caption{MAP at various cutoffs (1, 5, 10, 20, 50, 100) for selected pointwise reranking methods on FutureQueryEval, including InRanker Small, Transformer Ranker (electra-base), FlashRank (MiniLM-L-12-v2), MonoBERT (Large), Twolar (XL), and MonoT5 (T5 3B).}
    \label{fig:map2_pointwise}
\end{figure}

\paragraph{Pointwise Reranking (Table~\ref{tab:pointwise_our_app}).}
MonoT5-3B-10k emerges as the top-performing pointwise model with NDCG@10 of 60.75 and strong MAP scores across cutoffs (see Figure~\ref{fig:map2_pointwise}). Twolar-xl follows closely, showing robust fine-tuned performance on novel topics. FlashRank’s MiniLM-L-12-v2 model (55.43) and ColBERT-v2 (54.20) strike a compelling balance between ranking accuracy and efficiency, making them practical for real-time systems. Older UPR variants (e.g., GPT2, T5-small) lag significantly, indicating difficulty handling novel, unseen topics without task-specific tuning. This pattern holds across both top-k and full-range MAP metrics.

\begin{figure}[htbp]
    \centering
    \includegraphics[width=0.5\textwidth]{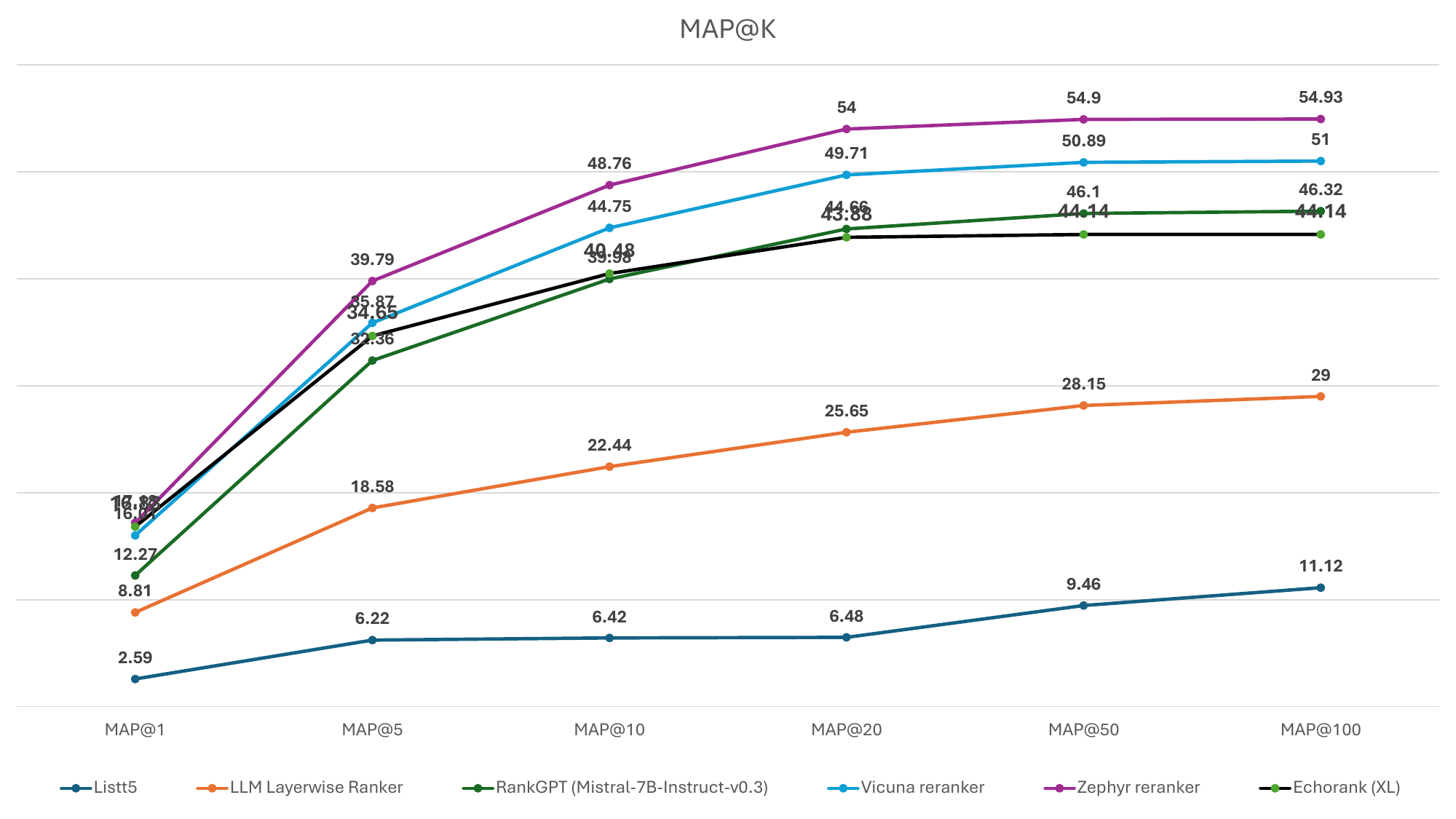}
    \caption{MAP at various cutoffs (1, 5, 10, 20, 50, 100) for selected listwise reranking methods on FutureQueryEval, including ListT5, LLM Layerwise Ranker, RankGPT (Mistral-7B-Instruct-v0.3), Vicuna Reranker, and Zephyr Reranker.}
    \label{fig:map3_listwise}
\end{figure}

\paragraph{Listwise Reranking (Table~\ref{tab:listwise_our_app}).}
Listwise models like Zephyr-7B and Vicuna-7B stand out, achieving NDCG@10 of 62.65 and 58.63 respectively, with Zephyr also leading MAP@10 (48.76, Figure~\ref{fig:map3_listwise}). These results validate the strength of generative rerankers that model full document lists. In contrast, ListT5-3B and base variants show very weak performance (NDCG@10 < 12), possibly due to outdated training data or lack of robustness to recent events. RankGPT variants using Mistral and LLaMA also perform competitively but remain a few points behind the top models. InContext rerankers provide moderate gains but do not match Zephyr or Vicuna.

\paragraph{Pairwise Reranking (Table~\ref{tab:pairwise_our}).}
EchoRank significantly improves over earlier versions, with Flan-T5-large and Flan-T5-XL reaching NDCG@10 scores of 54.8 and 54.97 respectively. These results rival top pointwise models, showing that well-optimized pairwise methods can generalize well to FutureQueryEval’s diverse and temporally fresh queries. However, their computational cost—due to pairwise comparisons across document sets—remains a bottleneck for scalability.

\paragraph{Insights.}
Across all types, reranking models trained or adapted to IR tasks (e.g., MonoT5, Zephyr) clearly outperform general-purpose or small zero-shot models (e.g., GPT2, InRanker-small). While listwise methods lead in overall accuracy, pointwise models provide efficient alternatives. FutureQueryEval thus offers a valuable diagnostic for evaluating models on truly unseen content, revealing performance gaps that traditional benchmarks may miss.

\begin{figure}[htbp]
    \centering
    \includegraphics[width=0.5\textwidth]{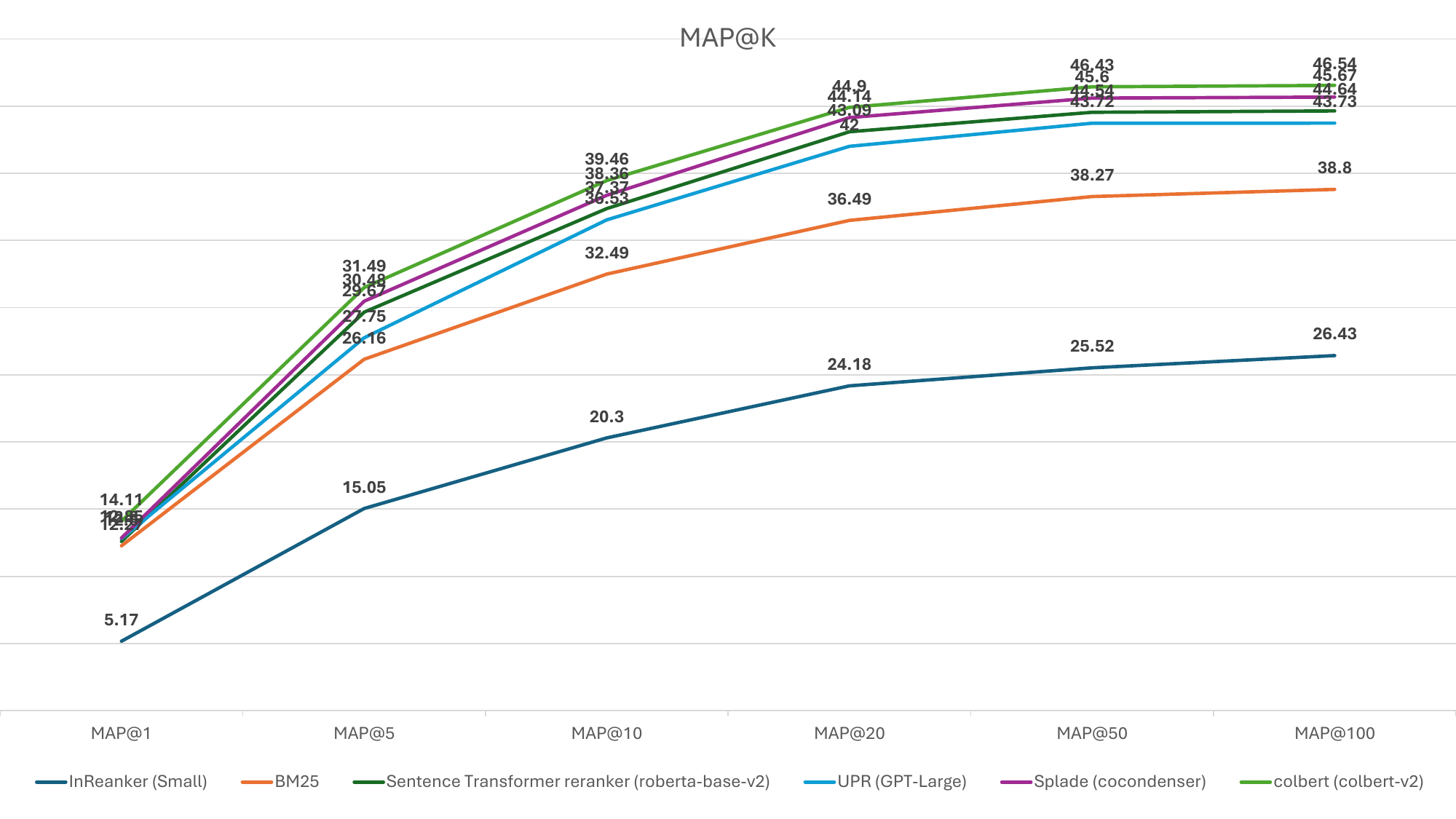}
    \caption{MAP at various cutoffs (1, 5, 10, 20, 50, 100) for selected pointwise reranking methods on FutureQueryEval, including InRanker Small, BM25, Sentence Transformer Reranker (roberta-base-v2), UPR (GPT-Large), Splade (cocondenser), and Colbert (colbert-v2).}
    \label{fig:map1_pointwise}
\end{figure}

\begin{table}[]
    \centering
\setlength\tabcolsep{2pt}
\resizebox{.95\linewidth}{!}{%
\begin{tabular}{ll|rrrrrr} 
\toprule
Method & Model & NDCG@1 & NDCG@5 & NDCG@10 & NDCG@20 & NDCG@50 & NDCG@100 \\
\midrule
BM25 & - & 39.86 & 43.01 & 46.42 & 50.26 & 53.79 & 55.44 \\

\midrule
\multirow{7}{*}{upr} & t5-small & 39.19 & 42.15 & 47.24 & 52.62 & 55.71 & 56.01 \\
 & t5-base & 42.57 & 45.05 & 49.67 & 54.89 & 57.57 & 57.78 \\
 & t5-large & 42.91 & 45.54 & 50.24 & 55.12 & 58.16 & 58.25 \\
 & t0-3b & 44.93 & 47.98 & 52.16 & 56.82 & 59.43 & 59.58 \\
 & gpt2 & 42.91 & 43.9 & 49.2 & 54.14 & 57.25 & 57.4 \\
 & gpt2-medium & 40.2 & 43.52 & 48.78 & 54.15 & 57.02 & 57.16 \\
 & gpt2-large & 43.58 & 44.98 & 50.3 & 55.18 & 58.12 & 58.18 \\

  \midrule
\multirow{5}{*}{flashrank} & ms-marco-TinyBERT-L-2-v2 & 45.61 & 45.89 & 50.29 & 53.97 & 57.6 & 58.39 \\
 & MiniLM-L-12-v2 & 53.04 & 51.72 & 55.43 & 59.64 & 61.95 & 62.21 \\
 & MultiBERT-L-12 & 20.61 & 23.3 & 26.91 & 31.07 & 37.32 & 41.87 \\
 & rank-T5-flan & 3.38 & 4.14 & 5.19 & 7.86 & 15.46 & 26.27 \\
 & MiniLM-L12-v2 & 51.01 & 51.47 & 54.64 & 58.44 & 61.22 & 61.49 \\
  \midrule
\multirow{8}{*}{monot5} & base & 53.04 & 55.15 & 57.88 & 61.61 & 63.58 & 63.7 \\
 & base-10k & 53.04 & 55.02 & 57.92 & 61.84 & 63.59 & 63.85 \\
 & large & 52.03 & 57.05 & 59.07 & 62.66 & 64.24 & 64.42 \\
 & mt5-base & 50.68 & 51.68 & 55.91 & 60.01 & 62.03 & 62.14 \\
 & mt5-base-v2 & 48.65 & 51.88 & 55.46 & 59.74 & 61.71 & 61.98\\
 & mt5-base-v1 & 51.35 & 52.26 & 55.24 & 59.62 & 61.88 & 62.12 \\
 & 3B-10k & 57.09 & 58.27 & 60.75 & 64.01 & 65.54 & 65.71 \\
  
  \midrule
\multirow{3}{*}{inranker} & inranker-small & 23.65 & 29.27 & 33.79 & 39.29 & 42.12 & 45.24 \\
 & inranker-base & 18.24 & 25.08 & 31.14 & 38.01 & 40.85 & 43.93 \\
 & inranker-3b & 18.24 & 28.47 & 32.39 & 38.11 & 41.22 & 44.5 \\
  \midrule
\multirow{10}{*}{transformer ranker} & mMiniLMv2-L12-H384-v1 & 51.35 & 51.96 & 54.9 & 58.89 & 61.51 & 61.81 \\
 & MiniLM-L-12-v2 & 52.03 & 51.01 & 54.99 & 59.26 & 61.55 & 61.8 \\
 & MiniLM-L-6-v2 & 53.04 & 51.94 & 55.51 & 59.55 & 61.95 & 62.23 \\
 & MiniLM-L-4-v2 & 46.96 & 49.77 & 53.29 & 57.33 & 60.07 & 60.39 \\
 & MiniLM-L-2-v2 & 40.88 & 45.33 & 50.02 & 54.31 & 57.37 & 57.9 \\
 & TinyBERT-L-2-v2 & 44.93 & 45.41 & 50.17 & 53.9 & 57.42 & 58.23 \\
 & electra-base & 48.99 & 52.24 & 55.56 & 59.61 & 61.55 & 62.05 \\
 & TinyBERT-L-6 & 46.28 & 50.21 & 53.58 & 57.34 & 59.79 & 60.37 \\
 & TinyBERT-L-4 & 47.3 & 48.67 & 51.91 & 55.71 & 58.26 & 59.46 \\
 & TinyBERT-L-2 & 42.23 & 43.6 & 47.94 & 52.25 & 55.88 & 56.85 \\
 
  \midrule
\multirow{1}{*}{Splade reranker} & splade-cocondenser & 46.96 & 49.02 & 52.93 & 57.72 & 60.08 & 60.21 \\
 \midrule
\multirow{9}{*}{\makecell[l]{Sentence \\ Transformer \\ Reranker}} & all-MiniLM-L6-v2 & 38.51 & 44.44 & 49.55 & 54.67 & 56.99 & 57.21 \\
 & gtr-t5-base & 44.26 & 47.82 & 52.25 & 56.8 & 59.15 & 59.33 \\
 & gtr-t5-large & 45.95 & 49.43 & 52.97 & 57.7 & 60.08 & 60.27 \\
 & gtr-t5-xl & 42.57 & 48.87 & 52.68 & 57.16 & 59.29 & 59.41 \\
 & sentence-t5-base & 42.91 & 45.34 & 50.29 & 54.19 & 57.58 & 57.85 \\
 & sentence-t5-xl & 43.92 & 46.69 & 51.2 & 55.86 & 58.28 & 58.63 \\
 & sentence-t5-large & 43.24 & 46.29 & 50.47 & 55.26 & 58.22 & 58.46 \\
 & bert-co-condensor & 31.76 & 37.96 & 42.4 & 48.03 & 52.25 & 52.82 \\
 & roberta-base-v2 & 44.26 & 46.94 & 51.26 & 56.12 & 58.53 & 58.91 \\
  \midrule
\multirow{2}{*}{colbert ranker} & colbert-v2 & 50.34 & 50.54 & 54.2 & 58.65 & 61.05 & 61.4 \\
 & mxbai-colbert-large-v1 & 46.96 & 50.08 & 53.96 & 58.18 & 60.57 & 60.91 \\
  \midrule
\multirow{1}{*}{monobert} & monobert-large & 50.0 & 53.39 & 56.99 & 61.16 & 63.03 & 63.11 \\
 \midrule
\multirow{3}{*}{llm2vec} & Meta-Llama-31-8B & 34.8 & 39.7 & 44.07 & 50.37 & 53.42 & 53.89 \\
 & Meta-Llama-3-8B & 32.77 & 37.53 & 42.84 & 48.87 & 52.44 & 52.98 \\
 & Mistral-7B-Instruct-v0.2  & 45.61 & 48.69 & 53.64 & 58.19 & 60.4 & 60.56 \\
  \midrule
\multirow{1}{*}{twolar} & twolar-xl & 55.41 & 57.79 & 60.03 & 63.68 & 65.19 & 65.23 \\

\bottomrule
\end{tabular} }
    \caption{ Performance of pointwise reranking methods on the FutureQueryEval dataset. Metrics reported include NDCG  at different cutoffs (1, 5, 10, 20, 50, and 100).}
    \label{tab:pointwise_our_app}
\end{table}

\begin{table}[]
    \centering
\setlength\tabcolsep{2pt}
\resizebox{0.95\linewidth}{!}{%
\begin{tabular}{ll|rrrrrr}
\toprule
Method & Model & NDCG@1 & NDCG@5 & NDCG@10 & NDCG@20 & NDCG@50 & NDCG@100 \\
\midrule
\multirow{2}{*}{listt5} & listt5-base & 12.84 & 13.28 & 11.5 & 11.12 & 25.39 & 30.83 \\
 & listt5-3b & 10.14 & 11.3 & 9.72 & 9.38 & 24.12 & 29.53 \\
  \midrule
\multirow{1}{*}{incontext reranker} & Mistral-7B-Instruct-v0.2 & 14.86 & 21.46 & 22.92 & 27.46 & 35.9 & 40.04 \\
 \midrule
\multirow{3}{*}{vicuna reranker} & rank vicuna 7b v1 & 54.05 & 55.9 & 59.09 & 62.31 & 64.03 & 64.44 \\
 & rank vicuna 7b v1 noda & 56.08 & 55.12 & 58.63 & 62.68 & 64.24 & 64.6 \\
  \midrule
\multirow{1}{*}{zephyr reranker} & rank zephyr 7b v1 full & 59.46 & 60.5 & 62.65 & 65.57 & 67.0 & 67.14 \\
\midrule
\multirow{5}{*}{rankgpt} & Llama-3.2-1B & 39.86 & 35.48 & 41.81 & 47.0 & 50.75 & 52.65 \\
 & Llama-3.2-3B & 39.86 & 50.36 & 52.79 & 56.22 & 58.85 & 59.55 \\
 & llamav3.1-8b & 39.86 & 52.19 & 54.18 & 57.71 & 60.15 & 60.66 \\
 & Mistral-7B-Instruct-v0.2 & 39.86 & 49.11 & 52.11 & 55.35 & 58.28 & 59.06 \\
 & Mistral-7B-Instruct-v0.3 & 39.86 & 53.34 & 55.21 & 58.47 & 60.62 & 61.25 \\
 \midrule
\multirow{2}{*}{llm layerwise ranker} 
 & bge-reranker-v2-gemma & 33.11 & 22.65 & 23.43 & 27.62 & 35.33 & 40.59 \\
 & bge-reranker-v2.5-gemma2 & 32.77 & 32.42 & 34.85 & 39.15 & 44.99 & 47.96 \\
   \midrule
\end{tabular} }

    \caption{Performance of listwise reranking methods evaluated on the FutureQueryEval dataset.}
    \label{tab:listwise_our_app}
\end{table}

\section{Comprehensive Pointwise Reranking Model Comparison}
\label{apndeix:pointwise}

This section presents a detailed comparison of pointwise reranking models across TREC DL and BEIR datasets (see Table~\ref{tab:pointwise_app_2}). The models vary significantly in architecture (e.g., BERT-based, T5-based, Condenser-based), size (from TinyBERT to 3B-scale models), and training setup (zero-shot, supervised, distilled).  Among all pointwise rerankers, \textbf{Inranker-3B} and \textbf{TWOLAR-XL} consistently achieve top-tier performance across nearly all domains, particularly excelling in scientific (SciFact: 78.31) and medical (Covid: 81.75) datasets. \textbf{RankT5-3B} and \textbf{MonoT5-3B} also perform strongly, especially in web-style queries (DL19, Robust04), showing the benefits of T5's encoder-decoder architecture. Lightweight models like \textbf{FlashRank-MiniLM} and \textbf{Transformer-Ranker-Base} provide a favorable trade-off between efficiency and performance, reaching over 70 nDCG@10 on DL19 and 65+ on BEIR tasks like DBPedia and News. Zero-shot models such as \textbf{UPR} (e.g., T5-Large or T0-3B) perform well in general-domain tasks but show weaker performance in niche domains like Touche or Signal. Larger models (3B+) show consistent improvements, particularly on complex or nuanced queries (e.g., Touche, SciFact), but some base-size models (e.g., RankT5-base, MonoT5-Large) still remain competitive, especially when fine-tuned.  Pointwise reranking methods scale well with model size and benefit from task-specific fine-tuning. Inranker-3B and TWOLAR-XL stand out as high-performing models, while FlashRank and MonoT5 variants offer robust, scalable alternatives. For general-purpose reranking with low latency, FlashRank-MiniLM and monoBERT remain strong choices.

\begin{table*}[!t]
\centering
\small
\setlength\tabcolsep{2pt}
\resizebox{1\linewidth}{!}{%

\begin{tabular}{l l cc|cccccccc}
\toprule
\textbf{Method} & \textbf{Model} & \textbf{DL19} & \textbf{DL20} & \textbf{Covid} & \textbf{NFCorpus} & \textbf{Touche} & \textbf{DBPedia} & \textbf{SciFact} & \textbf{Signal} & \textbf{News} & \textbf{Robust04} \\
\midrule
\multirow{7}{*}{UPR} & T5-Small & 49.94 & 47.07 & 62.81 & 30.46 & 21.54 & 31.98 & 61.68 & 29.71 & 27.41 & 33.14 \\
 & T5-Base & 55.12 & 53.94 & 64.12 & 31.12 & 19.42 & 34.28 & 65.56 & 30.50 & 27.32 & 33.40 \\
 & T5-Large & 58.33 & 56.05 & 68.69 & 31.71 & 18.94 & 35.35 & 65.44 & 32.80 & 25.25 & 34.48 \\
 
 & T0-3B & 60.18 & 59.55 & 68.83 & 33.48 & 23.97 & 34.41 & 71.21 & 33.02 & 39.10 & 41.74 \\
 & FLAN-T5-XL & 53.85 &  56.02 &  68.11 &  35.04  & 19.69  & 30.91  & 72.69  & 31.91 &  43.11 &  42.43  \\
 & GPT2 & 50.71 & 44.98 & 62.31 & 31.87 & 18.24 & 29.11 & 64.95 & 32.31 & 31.31 & 34.27 \\
 & GPT2-medium & 51.70 & 49.62 & 63.72 & 31.93 & 16.59 & 30.11 & 65.11 & 32.02 & 32.08 & 35.31 \\
 & GPT2-large & 52.48 & 0.467 & 63.23 & 33.62 & 16.72 & 30.76 & 32.49 & 66.59 & 30.85&  34.69\\
 \midrule

\multirow{5}{*}{FlashRank} & TinyBERT & 67.68 & 60.65 & 61.48 & 32.85 & 32.72 & 36.04 & 64.08 & 31.85 & 37.53 & 41.37 \\
 & MiniLM\footnote{\url{https://huggingface.co/cross-encoder/ms-marco-MiniLM-L12-v2}} & 70.80 & 66.27 & 69.06 & 33.02 & 34.77 & 42.77 & 66.28 & 33.62 & 44.54 & 47.18 \\
 & MultiBERT & 31.29 & 28.47 & 39.62 & 26.84 & 25.17 & 17.56 & 29.14 & 17.39 & 23.13 & 23.09 \\
 & T5-Flan & 21.79 & 17.02 & 38.61 & 18.11 & 8.23 & 7.77 & 8.29 & 6.57 & 12.06 & 15.40 \\
 & MiniLM\footnote{Fine-tuned on Amazon ESCI dataset} & 70.40 & 65.60 & 69.66 & 32.80 & 34.61 & 39.75 & 59.14 & 28.10 & 41.44 & 46.09 \\
 \midrule
\multirow{6}{*}{MonoT5} & Base & 70.81 & 67.21 & 72.24 & 34.81 & 38.24 & 42.01 & 73.14 & 30.47 & 45.12 & 51.24 \\
 & Base-10k & 71.38 & 66.31 & 74.61 & 35.69 & 37.86 & 42.09 & 73.39 & 32.14 & 46.09 & 51.69 \\
 & Large & 72.12 & 67.11 & 77.38 & 36.91 & 38.31 & 41.55 & 73.67 & 33.17 & 47.54 & 56.12 \\
 & Large-10k & 72.12 & 67.11 & 77.38 & 36.91 & 38.31 & 41.55 & 73.67 & 33.17 & 47.54 & 56.12 \\

 & mT5-Base & 70.81 & 64.77 & 73.77 & 34.36 & 35.62 & 40.11 & 71.17 & 29.79 & 45.34 & 48.99 \\

& 3B & 71.83 & 68.89 & 80.71 &  38.97 & 32.41 &44.45 &  76.57 & 32.55& 48.49 & 56.71\\
 \midrule
\multirow{3}{*}{RankT5} & T5-base & 72.13 & 67.91 & 75.63 & 34.99 & 41.24 & 42.39 & 73.37 & 30.86 & 44.07 & 52.19 \\
 & T5-large & 72.82 & 67.37 & 75.45 & 36.27 & 39.34 & 42.90 & 74.84 & 32.53 & 46.81 & 54.48 \\
 & T5-3b & 71.09 & 68.67 & 80.43 & 37.43 & 40.41 & 42.69 & 76.58 & 31.77 & 48.05 & 55.91 \\
  \midrule

\multirow{3}{*}{Inranker} & Inranker-small & 69.81 &61. 68 & 77.75 & 35.47 &  28.83 &  44.51  &74.90 & 29.37 & 46.29 & 50.91\\
 & Inranker-base  & 71.84 & 66.30& 79.84  &  36.58 & 28.97  & 46.50& 76.18 & 30.46 & 47.88 &  54.27\\
 & Inranker-3b  & 72.71 & 67.09 & 81.75  & 38.25  &29.24   &47.62& 78.31 &  32.20 & 49.63 &  62.47 \\
  \midrule
\multirow{10}{*}{Transformer Ranker} & mxbai-rerank-xsmall & 68.95 & 63.11 & 80.80 & 34.44 & 39.44 & 42.5 & 68.73 & 29.40 & 53.00 & 53.87 \\

 & mxbai-rerank-base & 72.49 & 67.15 & 84.00 & 35.64 & 34.32 & 42.50 & 72.33 & 30.20 & 51.92 & 55.59 \\
 
 & mxbai-rerank-large & 71.53 & 69.45 & 85.33 & 37.08 & 36.90 & 44.51 & 75.10 & 31.90 & 51.90 & 58.67 \\
 & bge-reranker-base & 71.17 & 66.54 & 67.50 & 31.10 & 34.30 & 41.50 & 70.60 & 28.40 & 39.50 & 42.90 \\
 & bge-reranker-large & 72.16 & 66.16 & 74.30 & 34.80 & 35.60 & 43.70 & 74.10 & 30.50 & 43.40 & 49.90 \\
 & bge-reranker-v2-m3 & 72.19 & 66.98 & 74.79 & 33.84 & 39.85 & 41.93 & 73.48 & 31.36 & 45.84 & 48.44 \\

 & bce-reranker-base & 70.45 & 64.13 & 67.59 & 33.90 & 27.50 & 38.14 & 70.15 & 27.31 & 40.48 & 48.13 \\
   
 & jina-reranker-tiny & 70.43 & 65.31 &  77.15 & 37.24 & 31.04 & 42.14 & 73.42 & 32.25 & 42.27 & 47.41\\

 & jina-reranker-turbo & 70.35 & 63.62 & 77.97 & 37.29 & 30.80 & 41.75 & 74.53 & 28.46 & 42.79 & 44.19 \\

  \midrule

\multirow{1}{*}{Splade Reranker} & Splade Cocondenser & 71.47 & 66.18 & 68.87 & 34.95 & 37.96 & 41.25 & 68.72 & 32.27 & 43.28 & 47.51 \\
 \midrule
\multirow{12}{*}{Sentence Transformer Reranker} & all-MiniLM & 63.84 & 60.40 & 70.83 & 33.10 & 29.23 & 34.87 & 65.63 & 28.50 & 45.42 & 46.03 \\
 & GTR-T5-base & 68.09 & 62.40 & 70.10 & 32.02 & 32.70 & 36.20 & 60.23 & 30.79 & 43.24 & 45.38 \\
 & GTR-T5-large & 67.23 & 63.33 & 69.50 & 33.03 & 32.84 & 38.20 & 62.41 & 31.19 & 44.32 & 46.98 \\
 & GTR-T5-xl & 67.55 & 64.51 & 69.63 & 33.39 & 34.28 & 38.76 & 63.65 & 31.10 & 45.73 & 47.95 \\
 & GTR-T5-xxl & 68.53 & 64.07 & 72.70 & 34.02 & 36.77 & 39.90 & 65.62 & 31.37 & 47.01 & 49.67 \\
 & sentence-T5-base & 51.15 & 49.37 & 66.02 & 30.17 & 24.63 & 33.67 & 47.29 & 29.78 & 41.71 & 48.24 \\
 & sentence-T5-xl & 54.95 & 53.22 & 67.01 & 31.72 & 29.78 & 36.38 & 50.73 & 31.22 & 43.58 & 48.33 \\
 & sentence-T5-xxl & 60.61 & 58.37 & 72.55 & 34.76 & 30.88 & 40.52 & 60.23 & 31.05 & 49.51 & 52.45 \\
 & sentence-T5-large & 55.36 & 54.20 & 63.57 & 30.23 & 28.08 & 31.89 & 47.40 & 30.56 & 42.94 & 47.06 \\
 & msmarco-bert-co-condensor & 56.34 & 53.50 & 62.20 & 28.38 & 20.12 & 31.93 & 53.04 & 31.16 & 36.56 & 36.99 \\
 & msmarco-roberta-base-v2 & 68.35 & 62.61 & 66.67 & 30.10 & 31.98 & 32.62 & 56.65 & 29.77 & 46.14 & 43.94 \\
  \midrule
\multirow{1}{*}{colbert ranker} & colbert-v2 & 69.02 & 66.78 & 72.6 & 33.70 & 35.51 & 45.20 & 67.74 & 33.01 & 41.21 & 45.83 \\
\midrule

monoBERT& BERT (340M)  & 70.50& 67.28 & 70.01& 36.88& 31.75& 41.87& 71.36 &31.44 &44.62& 49.35 \\
\midrule
Cohere Rerank-v2& - & 73.22& 67.08& 81.81 &36.36 & 32.51 & 42.51 & 74.44 & 29.60 &47.59 &50.78 \\
\midrule

Promptagator++ & - & &  &76.2 & 37.0  &38.1 & 43.4  & 73.1 &  & &  \\
\midrule
TWOLAR  & TWOLAR-Large & 72.82  & 67.61& 84.30&35.70  &33.4 & 47.8 & 75.6& 33.9&52.7  &58.3  \\
 & TWOLAR-XL &73.51  &  70.84   &82.70 & 36.60 &37.1 & 48.0 &76.5 & 33.8&50.8  & 57.9 \\

\bottomrule
\end{tabular} }
\caption{Performance comparison (nDCG@10) of various pointwise reranking models across standard TREC Deep Learning (DL19, DL20) and multiple BEIR benchmark datasets.}
\label{tab:pointwise_app_2}
\end{table*}

\section{Comprehensive Listwise Reranking Model Comparison}
\label{apndeix:ListWise}

Table~\ref{tab:Listwise_app} reports nDCG@10 performance for listwise rerankers across DL19, DL20, and BEIR. These methods jointly consider multiple documents to model inter-document relationships, with varying reliance on prompt design, input order sensitivity, and model size. \textbf{RankGPT (GPT-4)} remains the best-performing listwise reranker (DL19: 75.59, Covid: 85.51), highlighting the strengths of large generative models in capturing fine-grained semantic ordering. However, distilled models such as \textbf{LiT5-Distill-XL} and \textbf{Zephyr-7B} perform competitively while being significantly more efficient. \textbf{ListT5-3B} balances performance and latency, scoring above 71 on DL19 and exceeding 84 on Covid. \textbf{LiT5-Score} models show weaker results compared to their distilled counterparts, indicating that listwise distillation is more effective than pointwise scoring with listwise outputs. \textbf{InContext-Mistral-7B} and \textbf{LLaMA-based RankGPTs} (1B–3B) struggle on several benchmarks (DL19: 47–59), likely due to prompt length limitations and decoding inconsistencies. These findings suggest that prompt format and window size play a significant role in generative listwise performance. Listwise reranking benefits most from large-scale pretrained LLMs or their distilled variants. While GPT-4 leads in absolute performance, models like Zephyr and ListT5 offer practical alternatives. Distilled architectures (e.g., LiT5-Distill) provide strong performance at reduced cost, making them suitable for scalable deployment.

\begin{table*}[!t]
\centering
\small
\setlength\tabcolsep{2pt}
\resizebox{0.8\linewidth}{!}{%

\begin{tabular}{l l cc|cccccccc}
\toprule
\textbf{Method} & \textbf{Model} & \textbf{DL19} & \textbf{DL20} & \textbf{Covid} & \textbf{NFCorpus} & \textbf{Touche} & \textbf{DBPedia} & \textbf{SciFact} & \textbf{Signal} & \textbf{News} & \textbf{Robust04} \\

\midrule

InContext & Mistral-7B & 59.2 & 53.6 & 63.9 &33.2  & -& 31.4 &  72.4 &  & &  \\
& Llama-3.18B & 55.7  & 51.9 & 72.8 & 34.7 & -& 35.3 & 76.1 &  & &  \\
\midrule
\multirow{5}{*}{RankGPT} & Llama-3.2-1B & 47.13 & 44.93 & 59.29 & 31.43 & 37.20 & 26.04 & 67.49 & 26.94 & 38.76 & 38.14 \\
 & Llama-3.2-3B & 58.05 & 53.33 & 68.18 & 31.86 & 37.23 & 36.14 & 67.32 & 32.75 & 42.49 & 44.83 \\

& gpt-3.5-turbo&  65.80 &62.91 &76.67& 35.62 &36.18 &44.47& 70.43 &32.12& 48.85& 50.62\\
&gpt-4 & 75.59& 70.56 &85.51 &38.47& 38.57 &47.12& 74.95& 34.40 &52.89 &57.55\\

&llama 3.1 8b &58.46 & 59.68 & 69.61 & 33.62 & 37.98  & 37.25 & 69.82 & 32.95 &43.90 &49.59 \\

\midrule
\multirow{2}{*}{ListT5} & listt5-base & 71.80  & 68.10 & 78.30 & 35.60 &  33.40 & 43.70 & 74.10 & 33.50 & 48.50 & 52.10\\
 & listt5-3b & 71.80 & 69.10 & 84.70 & 37.70 & 33.60 & 46.20 & 77.0 &  33.80 & 53.20 & 57.80 \\ 
  \midrule
\multirow{6}{*}{lit5dist} & LiT5-Distill-base & 72.46 & 67.91 & 70.48 & 32.60 & 33.69 & 42.78 & 56.35 & 34.16 & 41.53 & 44.32 \\
 & LiT5-Distill-large & 73.18 & 70.32 & 73.71 & 34.95 & 33.46 & 43.17 & 66.70 & 30.88 & 44.41 & 52.46 \\
 & LiT5-Distill-xl & 72.45 & 72.46 & 72.97 & 35.81 & 32.76 & 43.52 & 71.88 & 31.23 & 46.59 & 53.77 \\
 & LiT5-Distill-base-v2 & 71.63 & 69.13 & 70.53 & 34.23 & 34.25 & 43.18 & 67.24 & 33.28 & 45.25 & 48.00 \\
 & LiT5-Distill-large-v2 & 72.15 & 67.78 & 73.10 & 34.05 & 34.55 & 43.35 & 69.30 & 31.16 & 42.42 & 50.95 \\
 & LiT5-Distill-xl-v2 & 71.94 & 71.93 & 73.08 & 34.68 & 34.29 & 44.59 & 69.68 & 32.76 & 45.88 & 51.70 \\
  \midrule
\multirow{3}{*}{lit5score} & LiT5-Score-base & 68.59 & 66.04 & 66.47 & 32.72 & 32.84 & 36.49 & 57.52 & 24.01 & 41.44 & 45.12 \\
 & LiT5-Score-large & 71.01 & 66.43 & 69.84 & 33.64 & 30.71 & 37.85 & 62.48 & 24.81 & 43.35 & 47.42 \\
 & LiT5-Score-xl & 69.36 & 65.56 & 69.66 & 34.36 & 29.09 & 39.10 & 67.50 & 24.07 & 44.95 & 52.88 \\
  \midrule
\multirow{1}{*}{Vicuna Reranker} & Vicuna 7b  & 67.19 & 65.29 & 78.30 & 32.95 & 32.71 & 43.28 & 70.49 & 32.87 & 44.98 & 47.83 \\
 \midrule
\multirow{1}{*}{Zephyr Reranker} & Zephyr 7B & 74.22 & 70.21 & 80.70 & 36.58 & 31.12 & 43.18 & 75.13 & 31.96 & 48.95 & 54.20 \\
 \midrule
\end{tabular} }
\caption{Evaluation results (nDCG@10) of listwise reranking approaches on TREC Deep Learning (DL19, DL20) and selected BEIR benchmarks.}
\label{tab:Listwise_app}
\end{table*}

\section{Comprehensive Open Domain QA Comparison}
\label{apndeix:open-domain}

 Table~\ref{tab:opendomain_app} presents a detailed comparison of reranking models for open-domain QA using BM25-retrieved candidates on Natural Questions (NQ) and WebQuestions (WebQ). The rerankers are evaluated at Top-1, Top-10, and Top-50 retrieval accuracy. \textbf{RankT5-3B}, \textbf{Lit5-Distill-XL-v2}, and \textbf{Twolar-XL} consistently achieve top performance across both datasets. RankT5-3B reaches the highest Top-1 scores (47.17 for NQ, 40.40 for WebQ), while Twolar-XL and Lit5-Distill-XL-v2 offer competitive results with improved Top-10 and Top-50 retrievals. Among efficient models, \textbf{FlashRank (MiniLM)} and \textbf{MonoBERT-large} perform surprisingly well—matching larger models on Top-50 accuracy. \textbf{InContext rerankers}, despite using powerful LLMs like LLaMA-3.1-8B, fall short on Top-1 (e.g., 15.15 on NQ), likely due to weak supervision and lack of fine-tuning. Sentence Transformer-based rerankers such as \textbf{GTR-XXL} and \textbf{GTR-XL} maintain robust Top-50 accuracy, indicating their suitability for shallow-depth QA pipelines. The results highlight the importance of both model size and supervision. Large-scale sequence-to-sequence models (e.g., RankT5, LiT5) dominate Top-1 accuracy, while dense retrievers like FlashRank and GTR achieve strong Top-50 recall. Models like Twolar-XL and Lit5-XL-v2 offer a practical balance of effectiveness and scalability, outperforming even several LLM-based rerankers in open-domain QA.

\begin{table*}[!ht]

\centering
\resizebox{0.7\textwidth}{!}{
\begin{tabular}{@{}l |l| c c c  | c c c@{}}\toprule
Reranking/ &  Model  &  \multicolumn{3}{c}{NQ} & \multicolumn{3}{c}{WebQ}  \\

& & Top-1 & Top-10 & Top-50 & Top-1 & Top-10 & Top-50   \\
\midrule
BM25      & - & 23.46 & 56.32 & 74.57 & 19.54  & 53.44 & 72.34 \\

\midrule
 
\multirow{2}{*}{UPR}  

  & T0-3B & 35.42 & 67.56 & 76.75 & 32.48 & 64.17 & 73.67 \\

  &gpt-neo-2.7B &28.75  & 64.81  &  76.56 &24.75    & 59.64 &  72.63  \\

\midrule

 \multirow{1}{*}{RankGPT} & llamav3.1-8b & 41.55 & 66.17 & 75.42 & 38.77 & 62.69 & 73.12 \\
 
\midrule

 \multirow{2}{*}{FlashRank} 
  & TinyBERT-L-2-v2& 31.49 & 61.57 & 74.95 & 28.54 & 60.62 & 73.17 \\
  & ce-esci-MiniLM-L12-v2 & 34.70 & 64.81 & 76.17 & 31.84 & 62.54 & 73.47  \\
\midrule
  
  \multirow{1}{*}{RankT5}
  &3b & 47.17& 70.85 & 76.89 & 40.40 & 66.58 & 74.45  \\

\midrule

\multirow{1}{*}{Inranker}  
&3b &15.90 & 48.06 & 69.00 & 14.46 & 46.11 & 69.34 \\
\midrule

\multirow{1}{*}{LLM2Vec} 
&Meta-Llama-31-8B & 24.32 & 59.55 & 75.26  &  26.72 &60.48 & 73.47\\

\midrule

\multirow{1}{*}{MonoBert} 
& large & 39.05 & 67.89 & 76.56 & 34.99 &64.56 & 73.96 \\

\midrule

\multirow{1}{*}{Twolar} 
& twolar-xl & 46.84 & 70.22 & 76.86 &41.68 &67.07 &74.40 \\

\midrule

\multirow{1}{*}{Echorank} 
& flan-t5-large& 36.73 & 59.11 & 62.38 & 31.74 &58.75 &61.51 \\
&flan-t5-xl&   41.68 & 59.05 & 62.38  & 36.22 &57.18 &61.51\\

\midrule

\multirow{2}{*}{\makecell[l]{Incontext\\ Reranker}} &  \multirow{2}{*}{llamav3.1-8b} & \multirow{2}{*}{15.15} & \multirow{2}{*}{57.11}  & \multirow{2}{*}{76.48}  & \multirow{2}{*}{18.89}   & \multirow{2}{*}{52.16} &  \multirow{2}{*}{71.70} \\
&&&&&&&\\

\midrule

\multirow{6}{*}{Lit5}
&  LiT5-Distill-base  & 40.05 & 65.95 & 75.73  & 36.76 & 63.48 & 73.12 \\
&  LiT5-Distill-large  & 44.40 & 67.59 & 76.01   &  39.66  &64.56 &73.67\\
&  LiT5-Distill-xl  &  47.81 & 68.55 & 76.26  &  42.37  &65.55 &73.62 \\
& LiT5-Distill-base-v2   &  42.57 &  66.73 & 75.56  & 39.61 &64.22 &73.32\\
&  LiT5-Distill-large-v2  & 46.53 &  67.83 & 75.87  & 41.97 &65.64 &72.98 \\
&  LiT5-Distill-xl-v2  & 47.92 & 69.03 & 76.17  & 41.53 &65.69 &73.27  \\

\midrule

\multirow{10}{*}{ \makecell[l]{Sentence \\ Transformer \\ Reranker} }
& GTR-base & 39.41 &65.95 & 76.03  & 36.56 &64.32 &73.62  \\
& GTR-large & 40.63 &68.25 & 76.73  & 38.97 &65.30 &73.57 \\
& T5-base & 31.19 &63.60 & 76.06  & 29.77 &62.84 &73.52 \\
& T5-large & 30.80 &63.35 &76.37   & 30.51  &61.71 &73.37 \\
&   all-MiniLM-L6-v2 & 33.35  & 65.37  &  76.01 & 30.95   & 62.10  & 73.52 \\
& GTR-xl   & 41.55  & 67.78  & 76.81 &  38.92   &  66.04  & 74.01 \\
& GTR-xxl   & 42.93 &  68.55&  77.00 &39.41    & 65.89 & 74.01 \\
&  T5-xxl  & 38.89  &  67.78  &76.64   &  35.82  & 65.20  & 74.01 \\
&  Bert-co-condensor  & 30.96  & 61.91   &  75.20 &  32.43  &  62.20  & 73.08 \\
&  Roberta-base-v2  & 32.60  & 63.24  & 75.42  &  31.34  &  62.64 &  73.37 \\
\midrule

\end{tabular}
}
\caption{Performance of re-ranking methods on BM25-retrieved documents for NQ Test and WebQ Test. Results are reported in terms of Top-1, Top-5, Top-10, Top-20, and Top-50 accuracy, highlighting the impact of various re-ranking models on retrieval effectiveness. Please note that some results may differ from the original papers (e.g., UPR) as our experiments were conducted with the top 100 retrieved documents, whereas the original studies used 1,000 documents for ranking.
}
\label{tab:opendomain_app}
\end{table*}
\end{document}